\documentclass{article}

\usepackage[preprint]{neurips_2025}

\usepackage[utf8]{inputenc} 
\usepackage[T1]{fontenc}    
\usepackage{hyperref}       
\usepackage{url}            
\usepackage{booktabs}       
\usepackage{subcaption}     
\usepackage{amsfonts}       
\usepackage{nicefrac}       
\usepackage{microtype}      
\usepackage{xcolor}         
\usepackage{amsmath}
\usepackage{amssymb}
\usepackage{mathtools}
\usepackage{amsthm}
\usepackage{algorithm}
\usepackage{algorithmic}
\theoremstyle{plain}
\usepackage{microtype}
\usepackage{graphicx}
\usepackage{wrapfig}
\usepackage{enumitem}

\theoremstyle{definition}

\theoremstyle{remark}

\title{Bidirectional Soft Actor-Critic: Leveraging Forward and Reverse KL Divergence for Efficient Reinforcement Learning}

\author{
  Yixian Zhang, Huaze Tang, Changxu Wei, Wenbo Ding \\
  Tsinghua University
}

\begin{document}

\maketitle

\begin{abstract}
  The Soft Actor-Critic (SAC) algorithm, a state-of-the-art method in maximum entropy reinforcement learning, traditionally relies on minimizing reverse Kullback-Leibler (KL) divergence for policy updates. However, this approach leads to an intractable optimal projection policy, necessitating gradient-based approximations that can suffer from instability and poor sample efficiency. This paper investigates the alternative use of forward KL divergence within SAC. We demonstrate that for Gaussian policies, forward KL divergence yields an explicit optimal projection policy---corresponding to the mean and variance of the target Boltzmann distribution's action marginals. Building on the distinct advantages of both KL directions, we propose Bidirectional SAC, an algorithm that first initializes the policy using the explicit forward KL projection and then refines it by optimizing the reverse KL divergence. Comprehensive experiments on continuous control benchmarks show that Bidirectional SAC significantly outperforms standard SAC and other baselines, achieving up to a $30\%$ increase in episodic rewards, alongside enhanced sample efficiency.
\end{abstract}

\section{Introduction}
Maximum entropy reinforcement learning has emerged as a powerful framework for tackling complex decision-making problems, including mobile robot control \citep{Robot_Sac}, autonomous driving \citep{Driving_Sac}, and protein engineering \citep{protein_eng}. The Soft Actor-Critic (SAC) algorithm \citep{SAC_ICML}, a prominent algorithm within this framework, has demonstrated superior performance across a variety of continuous benchmark tasks. In SAC, the optimization of the policy is achieved by seeking a projection policy with respect to the Boltzmann distribution based on the Q-function, where the reverse Kullback-Leibler (KL) divergence is used as the probabilistic measure. However, due to the intractability of the projection policy, gradient-based methods are employed to iteratively adjust the parameterized policy towards the optimal projection \citep{SAC_arxiv}. Despite their utility, these optimization techniques can exhibit instability and require a substantial number of samples to achieve convergence \citep{F_R_KL,Sample_SAC1}. In fact, even in the simple case of a Gaussian distribution, we observe that gradient optimization methods often fail to converge.

The use of reverse KL divergence as a measurement to guide policy projection has two key advantages, as outlined in prior studies \citep{SAC_arxiv,SAC_recent1,SAC_recent2}: First, minimizing the reverse KL divergence ensures that the expected Q-function value of the updated policy is greater than that of the previous policy. Second, the gradient of the reverse KL divergence is computationally tractable, avoiding the need for the intractable normalization term in the Boltzmann distribution. However, if the reverse KL divergence optimization becomes unstable, such as when the KL divergence increases during gradient optimization, these advantages are lost. An alternative approach involves the forward KL divergence, which is commonly utilized in variational policy optimization \citep{F_KL1,F_KL2}, where few papers concentrate on its use in the SAC algorithm.

In this paper, we find that, by replacing the reverse KL divergence with the forward KL divergence in the SAC framework, denoted as Forward SAC, the optimal projection policy becomes explicit. Specifically, for Gaussian policies—widely used in stochastic continuous actor-critic algorithms \citep{SAC_ICML,Gaussian_actor1,Gaussian_actor2,Gaussian_actor3}—the optimal projection policy within forward KL divergence corresponds to the mean and variance of the action under the Boltzmann distribution. This insight allows us to directly compute the optimal projection policy using numerical integration, thereby achieving stable and sample-efficient policy updating. However, it is not guaranteed that the updated policy within forward KL divergence can maximize the expected Q value. These characteristics of the two directions of KL divergence lead to a direct idea: Can we integrate the advantages of these two directions of KL divergence? Therefore, we propose the Bidirectional SAC algorithm, which uses the optimal projection policy in forward KL divergence as the initialization policy, and it optimizes the policy by minimizing the reverse KL divergence. To efficiently compute the mean and variance, we adapt the Value Decomposition Networks (VDN) \citep{VDN}, originally designed for multi-agent reinforcement learning, into the single-agent context for Q-function learning, where we denote it as VDN-a networks. Our theoretical analysis, grounded in Copula theory, elucidates how the VDN-a architecture is designed to approximate the necessary marginal properties of the Boltzmann distribution. Supported by experimental results, this approach facilitates the efficient learning required for the independent numerical integration of the mean and variance for each action dimension. Comparative experiments in MuJoCo and Box2D environments \citep{MuJoCo,Box2D} demonstrate that the Bidirectional SAC algorithm outperforms the SAC algorithm and other baseline methods, achieving up to $30\%$ increase in episodic rewards.



Our contributions are as follows:
\begin{itemize}[topsep=-1pt,itemsep=-0.5pt,leftmargin=*]
    \item We introduce the Forward SAC algorithm, in which the updated policy is the optimal projection to the Boltzmann distribution of the Q-function. 
    \item We conduct a thorough analysis for the use of forward and reverse KL divergence in the SAC algorithm, where we find that the use of both directions has its advantages and disadvantages. To leverage these advantages, we propose the Bidirectional SAC algorithm, which optimizes the policy sample efficiently.
    \item To efficiently compute the optimal projection, we propose VDN-a networks for single-agent Q-function learning. Our theoretical analysis and experimental results demonstrate its efficacy in approximating the marginals of the Boltzmann distribution, crucial for efficiently computing the forward KL optimal projection within Bidirectional SAC. Our experiments demonstrate that the Bidirectional SAC algorithm converges rapidly and outperforms baseline algorithms, achieving up to $30\%$ increase in episodic rewards.
\end{itemize}

\section{Preliminaries and background}
In this section, we briefly introduce the maximum entropy reinforcement learning and the famous soft actor-critic (SAC) algorithm \citep{SAC_ICML}. 

\subsection{Maximum entropy reinforcement learning}
We consider policy search in an infinite-horizon Markov decision process $\langle\mathcal{S}, \mathcal{A}, p, r, \rho\rangle$, with continuous state $\mathcal{S}$ and action $\mathcal{A}$ space. The transition function $p: \mathcal{S} \times \mathcal{A} \times \mathcal{S} \rightarrow [0, \infty)$ defines the probability density of transitioning, and rewards are $r_t = r(s_t, a_t) \in [r_{\min}, r_{\max}]$. 

The objective of reinforcement learning is to learn an optimal policy $\pi^{*}$ that maximizes the cumulative expected rewards  $\pi^{*}=\arg \max _\pi  \mathbb{E}_{\tau \sim \pi} \left[\sum_{l=t} r_l\right]$, 
where $\tau$ denotes the trajectory under policy $\pi$ originating at $(s_t, a_t)$ \citep{Suttonbook}. The maximum entropy reinforcement learning extends the standard objective by incorporating an entropy term $\mathcal{H}(\pi(\cdot \mid s)) = - \mathbb{E}_{a \sim \pi(\cdot \mid s)} \left[ \log \pi(a \mid s) \right]$, thereby encouraging policies to maintain stochasticity and explore more effectively. In this setting, the optimal policy $\pi_{\text{Ent}}^{*}$ seeks to maximize both the expected cumulative rewards and the entropy of the policy at each state \citep{SAC_arxiv}:
\begin{equation}
    \label{entro_obej}
    \pi_{\text {Ent }}^*=\arg \max _\pi  \mathbb{E}_{\tau \sim \pi}\left[\sum_{l=t} r_l+\alpha \mathcal{H}\left(\pi\left(\cdot \mid s_l\right)\right)\right],
\end{equation}
where $\alpha$ is the temperature parameter to determine the relative importance of entropy and reward. 

\subsection{Soft actor-critic}
Equation \eqref{entro_obej} is addressed by various methods \citep{InverseRL,MPO,maximum_recent}. Soft Q-Learning (SQL) \citep{SQL}, an early Q-learning approach, used an approximate sampler for the actor, limited by posterior approximation accuracy. To overcome this limitation, the Soft Actor-Critic (SAC) algorithm is introduced \citep{SAC_ICML}, where the actor is parameterized within a predefined policy class. In the SAC algorithm, the soft Q-function under policy $\pi$ is defined as:
\vspace{-5pt}
\begin{equation}
    \nonumber
   Q^\pi(s_t, a_t) = r_t+\mathbb{E}_{\tau \sim \pi}\left[\sum_{l=t+1}\left(r_l+\alpha\mathcal{H}\left(\pi\left(\cdot \mid s_l\right)\right)\right)\right],
\end{equation}
 The policy in SAC is typically modeled using a parameterized family of distributions, often instantiated as a multivariate Gaussian distribution family $\Pi$ \citep{Gaussian_AAAI, Gaussian_nips}, to enable efficient exploration and exploitation. In the update process, the soft Q-function and Gaussian policy are typically parameterized by neural networks $Q_{\theta}(s_t, a_t)$ and $\pi_{\phi}(a_t | s_t)$, respectively, where $a_t \sim \mathcal{N}(f_{\phi}(s_t), \Sigma_{\phi})$. Parameters of these neural networks are denoted as $\theta$ and $\phi$. The soft Q-function is updated with \citep{SAC_arxiv}:
\begin{equation}
\label{Q_updated}
        \nabla_\theta J_Q(\theta) = \nabla_\theta Q_\theta\left(a_t,s_t\right) \left[ Q_\theta\left(s_t, a_t\right) - \left( r_t  +  Q_{\bar{\theta}}\left(s_{t+1}, a_{t+1}\right) - \alpha \log  \pi_\phi\left(a_{t+1} \mid s_{t+1}\right)   \right) \right],
\end{equation}
where $Q_{\bar{\theta}}(s_{t+1}, a_{t+1})$ represents the target soft Q-function with parameters $\bar{\theta}$, which are updated periodically to stabilize training \citep{Target_Q}. The policy is optimized by the soft policy improvement, which projects the Boltzmann distribution $q(\cdot \mid s_t) = \frac{\exp \left(\frac{1}{\alpha} Q^{\pi_{\text {old }}}\left(s_t, \cdot\right)\right)}{Z^{\pi_{\text {old }}}\left(s_t\right)}$ onto the Gaussian distribution family through minimization of the reverse Kullback-Leibler (KL) divergence $D_{\mathrm{KL}}(\pi||q)$:
\vspace{-5pt}
\begin{equation}
    \label{Projection_problem}
    \pi_{\text {new-r}}=\arg \min _{\pi^{\prime} \in \Pi} D_{\mathrm{KL}}\left(\pi^{\prime}\left(\cdot \mid s_t\right) \bigg\| \frac{\exp \left(\frac{1}{\alpha} Q^{\pi_{\text {old }}}\left(s_t, \cdot\right)\right)}{Z^{\pi_{\text {old }}}\left(s_t\right)}\right),
\end{equation}
where the partition function $Z^{\pi_{\text{old}}}(s_t) = \int_{\mathcal{A}} \exp\left( \frac{1}{\alpha} Q^{\pi_{\text{old}}}(s_t, a) \right) da$ ensures that the distribution is normalized. In practice, policy parameters are typically updated using stochastic gradient descent, where an approximated projection policy $\tilde{\pi}_{\text{new-r}}$ is computed as:    
\begin{equation}
\label{SAC_updating}
    \tilde{\pi}_{\text{new-r}} = \pi_{\text{old}} - \beta \nabla_\phi J_\pi(\phi),
\end{equation}
with the gradient update given by $\nabla_\phi J_\pi(\phi) = \nabla_\phi \left[\alpha \log \pi_\phi\left(a_t \mid s_t\right) - Q_\theta\left(s_t, a_t\right)\right]$, where $a_t$ is implicitly parameterized by $\phi$ through $a_t \sim \mathcal{N}(f_{\phi}(s_t), \Sigma_{\phi})$, and $\beta$ denotes the learning rate. This updating rule directs the new policy $\tilde{\pi}_{\text{new-r}}$ towards the true projection $\pi_{\text{new-r}}$, but the resulting policy often remains some distance away from the true projection itself. Nevertheless, this approach is widely used in prominent SAC-based algorithms, including conservative Q-Learning \citep{CQL}, inverse soft Q-Learning \citep{IQ-learn}, and distributional soft actor-critic \citep{Distribution_SAC}, among others. 
However, applying gradient descent to optimize the reverse KL divergence can suffer from instability and inefficiency, largely attributable to its pathological curvature \citep{F_R_KL}. Our theoretical analysis and empirical findings indicate that this optimization approach frequently yields optimized distributions that deviate significantly from the true target distribution, even when applied in relatively simple scenarios.

In contrast to the reverse KL divergence traditionally employed in SAC, the forward KL divergence $D_{\mathrm{KL}}(q||\pi')$ has been extensively investigated in alternative reinforcement learning frameworks, including variational policy optimization~\citep{log_like2, log_like3, log_like4}. These methodologies present a fundamentally different perspective on policy optimization and have demonstrated comparable or superior performance across various benchmark environments~\citep{inference, constraint_MPO}. Despite these promising results, a systematic examination of the utilization and comparative efficacy of different KL divergence directions within the SAC framework remains notably absent from the literature. Our work addresses this critical research gap by rigorously analyzing the standard SAC algorithm and methodically exploring the integration of forward KL divergence within this framework. This investigation ultimately leads to our proposed Bidirectional SAC algorithm, which strategically combines the complementary advantages of both KL divergence directions to enhance sample efficiency and asymptotic performance.




\section{Two directions of KL divergence in SAC algorithm}
\label{Analysis}

In this section, we show that the use of gradient descent for optimizing reverse KL divergence is unstable, where a single gradient step is hard to decrease the reverse KL divergence even for a simple target distribution. For the use of forward KL divergence in the SAC algorithm, we derive the explicit form of the projection policy in the forward direction of KL divergence, which can be efficiently computed through numerical integration. Ultimately, we introduce a practical way to efficiently estimate numerical integration by using specially designed neural networks. 


\subsection{Gradient descent in reverse KL divergence}
\begin{figure}[htbp]
    \centering
    \begin{subfigure}[b]{0.49\linewidth}
        \centering
        \includegraphics[width=\columnwidth]{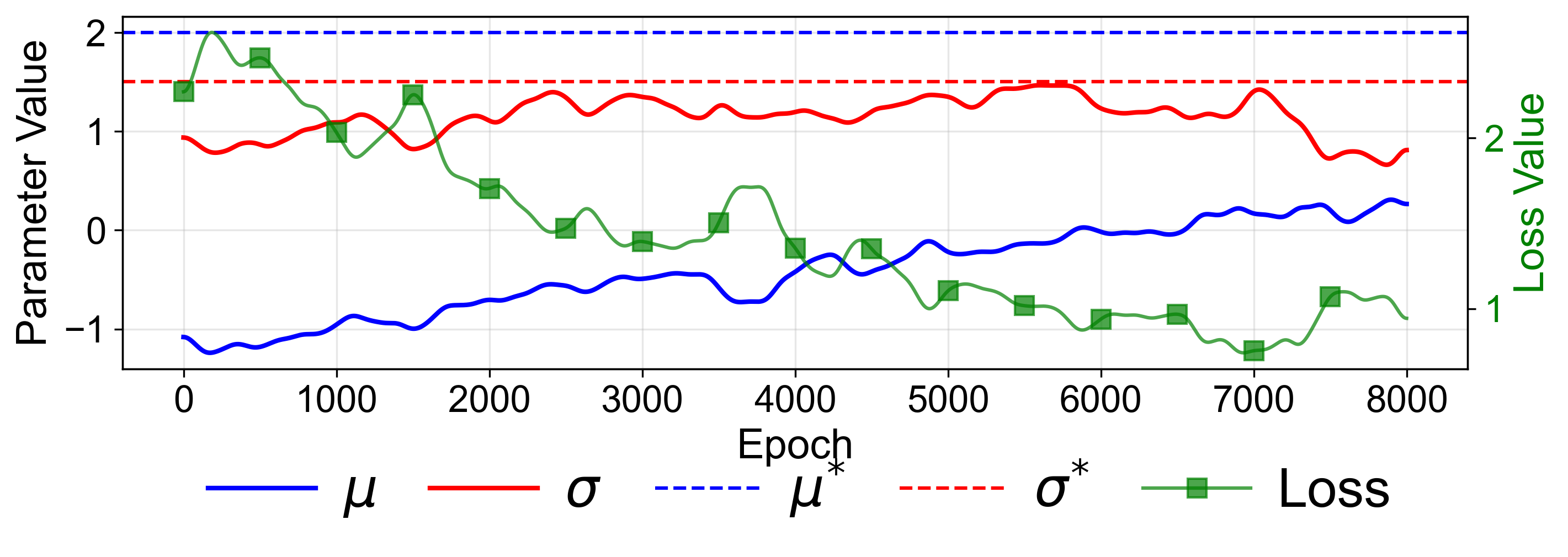}
        \caption{  }
        \label{conver}
    \end{subfigure}
    \hfill
    \begin{subfigure}[b]{0.49\linewidth}
        \centering
        \includegraphics[width=\columnwidth]{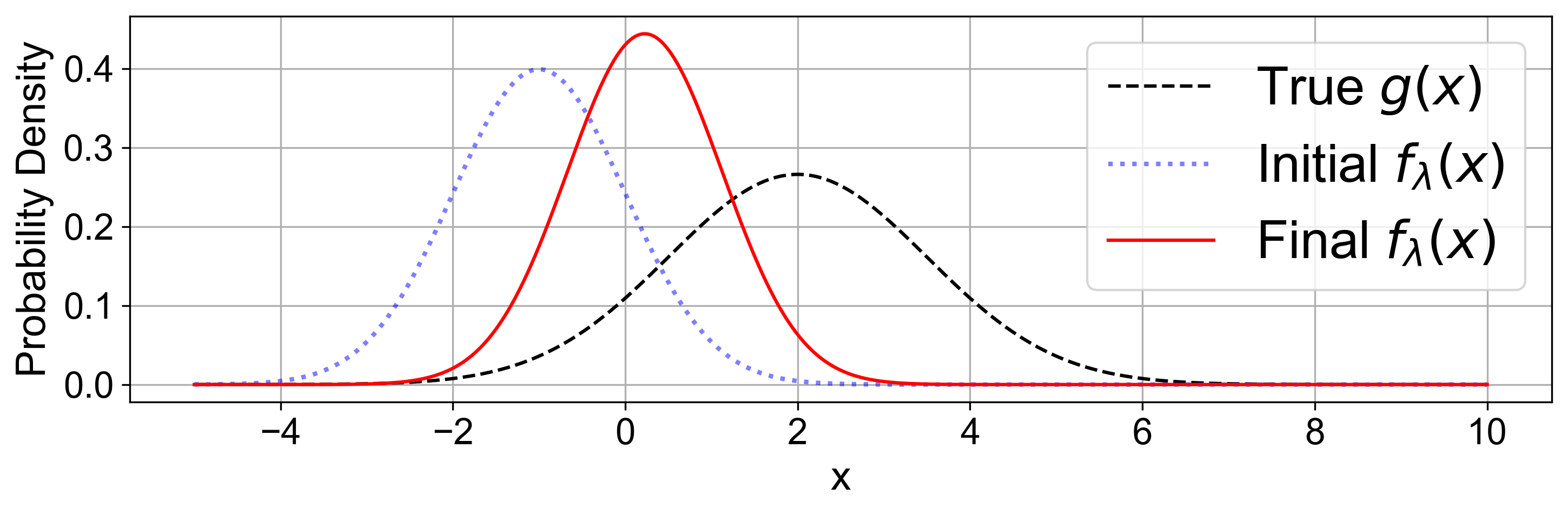}
        \caption{  }
        \label{distri}
    \end{subfigure}
    \caption{Optimization of the parameterized distribution $f_\lambda(x)$ towards the true distribution $g(x)$ using gradient descent with reverse KL divergence as the loss function. (a) The optimization process with respect to the parameters $\lambda = (\mu, \sigma)$ and the corresponding change in loss function. (b) Comparison of the initial and final distributions after optimization, alongside the true distribution.}
    \label{visualize_gradient}
    \vspace{-5pt}
\end{figure}

To examine the effectiveness of gradient-based methods in minimizing reverse KL divergence, we consider two Gaussian distributions: the true distribution $g(x)$ and a parameterized distribution $f_\lambda(x)$, where $\lambda = (\mu, \sigma)$ represents the mean and standard deviation. The loss function is the reverse KL divergence $D_{\mathrm{KL}}\left(f_\lambda(x) | g(x)\right)$, and we optimize the parameters $\lambda$ using gradient descent over 8000 epochs to approximate the true distribution. Let the true distribution $g(x)$ have mean $\mu^*$ and standard deviation $\sigma^*$. The optimization process is illustrated in Figure \ref{visualize_gradient}. 

The results show that while the gradient-based optimization method leads to a reduction in the KL divergence loss, it struggles to accurately converge to the optimal parameters $\mu^*$ and $\sigma^*$, as demonstrated in the distribution comparison on the right side of Figure \ref{visualize_gradient}. This behavior can be attributed to the inherent instability of the reverse KL divergence, which depends on samples drawn from the distribution to be optimized. Such instability has been shown to slow down the convergence process \citep{F_R_KL, F_R_KL_Els}. Therefore, methods that directly target the true distribution may offer more efficient learning. In the following subsection, we demonstrate that employing the forward KL divergence, which learns directly from the mean and variance of the true distribution, significantly enhances optimization efficiency.

\subsection{The explicit form of the projection policy}
Considering the forward KL divergence, reformulating Problem~\eqref{Projection_problem} as:
\begin{equation}
\label{Projection_problem_inv}
    \pi^{}_{\text {new-f}}=\arg \min _{\pi^{\prime} \in \Pi} D_{\mathrm{KL}}\left( q(\cdot \mid s_t) \bigg\| \pi^{\prime}\left(\cdot \mid s_t\right)  \right).
\end{equation}

This optimization problem can be resolved by this key observation: while the covariance matrix in a Gaussian policy $\pi^{\prime}$ is theoretically dense, it is commonly approximated as a diagonal matrix for computational simplicity in optimization \citep{SAC_ICML, stb3, cleanRL}. Consequently, the $n$-dimensional multivariate Gaussian distribution simplifies to $n$ independent one-dimensional Gaussian distributions, where the correlations between action dimensions are implicitly modeled by the neural networks.  
We begin by refining the optimization objective as follows:
\begin{equation}
\label{final_objective}
    D_{\mathrm{KL}}\left( q(\cdot \mid s_t)\bigg\| \pi^{\prime}(a \mid s_t)\right) = \int_{\mathcal{A}} q(a \mid s_t) \log q(a \mid s_t) - q(a \mid s_t) \log \pi^{\prime}(a \mid s_t) \, da.
\end{equation}
For convenience, we denote the forward KL divergence as $D_{\mathrm{KL}}\left( q(\cdot \mid s_t)\| \pi^{\prime}(a \mid s_t)\right)$ as $D_{\mathrm{KL}}$ for the remainder of this paper. Denoting the dimension of action space $\mathcal{A}$ is $N$, let $\pi^{\prime}(a \mid s_t) = \prod_i^N \pi_i^{\prime}(a^i \mid s_t)$, where each $\pi_i^{\prime}(\cdot \mid s_t) = \mathcal{N}(f_{\phi}^{i}(s_t), \Sigma_{\phi}^{i})$ represents a one-dimensional Gaussian distribution in dimension $i$, and $f_{\phi}^{i}(s_t), \Sigma_{\phi}^{i}$ represent the corresponding network output and the standard deviation of the action in dimension $i$, respectively.
 We derive the partial derivative with respect to $f_{\phi}^{i}(s_t)$ and $\Sigma_{\phi}^{i}$ in the objective function of Equation~\eqref{final_objective} as follows:
\begin{equation}
\nonumber
    \frac{\partial D_{\mathrm{KL}}}{\partial f_{\phi}^{i}(s_t)} = \frac{1}{\Sigma_\phi^i} \int_{\mathcal{A}_i} q_i(a^i) (a^i - f_{\phi}^{i}(s_t)) \, da^i, \quad \frac{\partial D_{\mathrm{KL}}}{\partial \Sigma^{i}_{\phi}}  =  \frac{1}{2 \Sigma_\phi^i} - \frac{1}{2 (\Sigma_\phi^i)^2} \int_{\mathcal{A}_i} q_i(a^i ) (a^i - f_\phi^i(s_t))^2 \, da^i.
\end{equation}
Here, $q_i(a^i ) = q_i(a^i \mid s_t)$ represents the marginal distribution of $q(a \mid s_t)$ with respect to dimension $i$. 
The detailed derivations are provided in Appendix \ref{KL_partial_diri}. By setting $\frac{\partial D_{\mathrm{KL}}}{\partial f_{\phi}^{i}(s_t)} = 0$ and $\frac{\partial D_{\mathrm{KL}}}{\partial \Sigma_{\phi}^{i}} = 0$, we can solve for the optimal solution of Problem~\eqref{Projection_problem_inv}, yielding the optimal $f^{i}(s_t)^{*}$ and $\Sigma^{i}{}^{*}$:
\begin{equation}
\nonumber
    f^{i}(s_t)^{*} =   \int_{\mathcal{A}_i} q_i(a^i \mid s_t) a^i \, da^i, \quad \Sigma^{i}{}^{*} = \int_{\mathcal{A}_i} q_i(a^i \mid s_t) (a^i - f_{\phi}^{i}(s_t))^2 \, da^i.
\end{equation}
Thus, the optimal solution to Problem~\eqref{Projection_problem_inv} is elegantly expressed in terms of the mean and variance of the marginal distribution of $q(\cdot \mid s_t)$, offering an explicit projection policy that does not rely on iterative optimization methods such as gradient-based approaches.

\subsection{VDN-a network for efficient numerical integration}

\begin{wrapfigure}{r}{0.45\textwidth} 
  \vspace{-10pt} 
  \centering
  \includegraphics[width=\linewidth]{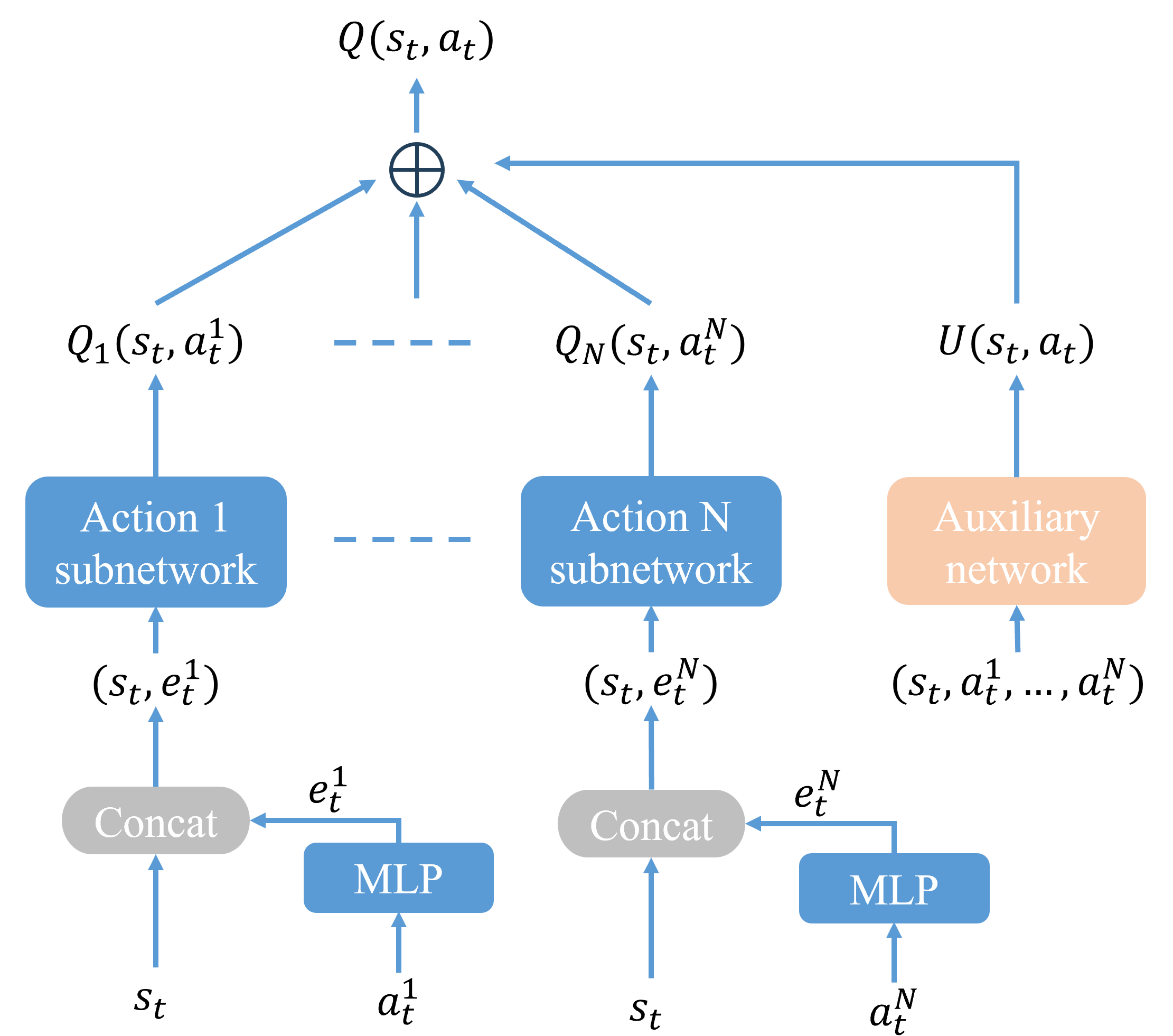} 
  \caption{The structure of VDN-a network}
  \label{fig:VDN-a network} 
  \vspace{-10pt} 
\end{wrapfigure}

To compute the optimal values $f^{i}(s_t)^{*}$ and $\Sigma^{i}{}^{*}$ efficiently, we introduce the VDN-a network as the critic network to learn the marginal Q values $Q_i(s_t, a_t^i)$ for each action dimension $a_t^i$. This approach is inspired by Value Decomposition Networks (VDN), where a brief introduction to it is provided in Appendix \ref{VDN_net}. Similar to VDN, which designs an independent subnetwork for each agent, the VDN-a network treats each action dimension along with the entire state as an individual "agent" and constructs an independent subnetwork for it. The architecture of the VDN-a network is illustrated in Figure \ref{fig:VDN-a network}. Since each action $a_t^i$ is a single-dimensional data point, it is mapped to an embedding $e_t^i$ through a multilayer perceptron (MLP) for complex pattern recognition. These embeddings are then concatenated with the state $s_t$ to form $(s_t, e_t^i)$, which are fed into the $i$-th subnetwork to obtain the marginal Q value $Q_i(s_t, a_t^i)$ for action $a_t^i$. To enhance the learning capacity of the VDN-a network, an auxiliary network is also incorporated. This auxiliary network takes the entire state and action tuple $(s_t, a_t^1, \dots, a_t^N)$ as input and outputs $U(s_t, a_t)$. The outputs of each subnetwork are aggregated by summing them together with the auxiliary network's output to form the final Q value $Q(s_t, a_t)$. 
Once the marginal Q values are learned, the marginal distributions $q_i(\cdot \mid s_t)$ are obtained by $
q_i(a_t^i \mid s_t) = \frac{ \exp \left( \frac{1}{\alpha} Q_{i}(s_t, a_t^i) \right) }{ \int_{\mathcal{A}_i} \exp \left( \frac{1}{\alpha} Q_{i}(s_t, a^i) \right) \, da^i },
$ where $\mathcal{A}_i$ is the $i$-th dimension of the action space and the integration is performed using numerical integration, as detailed in the Appendix \ref{Integration_details}. We also prove in the Appendix \ref{theoretic_copula} that if the auxiliary network is sufficiently learned or each dimension of the action is independent, the learned $q_i(a_t^i \mid s_t)$ is actually the true marginal distribution of the $q(\cdot \mid s_t)$. To compute the optimal values $f^{i}(s_t)^{*}$ and $\Sigma^{i}{}^{*}$, only the mean and variance of the marginal distribution $q_i(\cdot \mid s_t)$ are required. While direct estimation of the complete marginal distribution remains computationally intractable, empirical results demonstrate that the proposed critic network architecture effectively captures the first two moments—specifically the mean and variance—of these marginal distributions through numerical integration techniques. The mathematical formulation and implementation details of this integration procedure are presented in Appendix \ref{Integration_details} due to space constraints.

\section{The Bidirectional SAC}
This section provides a general analysis of the properties of the two directional variants of SAC algorithms. The analysis reveals that each directional variant possesses distinct advantages and disadvantages. Subsequently, we demonstrate that the Bidirectional SAC effectively integrates the strengths of both directional approaches while mitigating their respective limitations.
\subsection{Analysis of the use in forward and reverse KL divergence in SAC}

\begin{figure*}[htbp]
 \centering
 \begin{subfigure}[b]{0.49\linewidth}
 \centering
\includegraphics[width=\columnwidth]{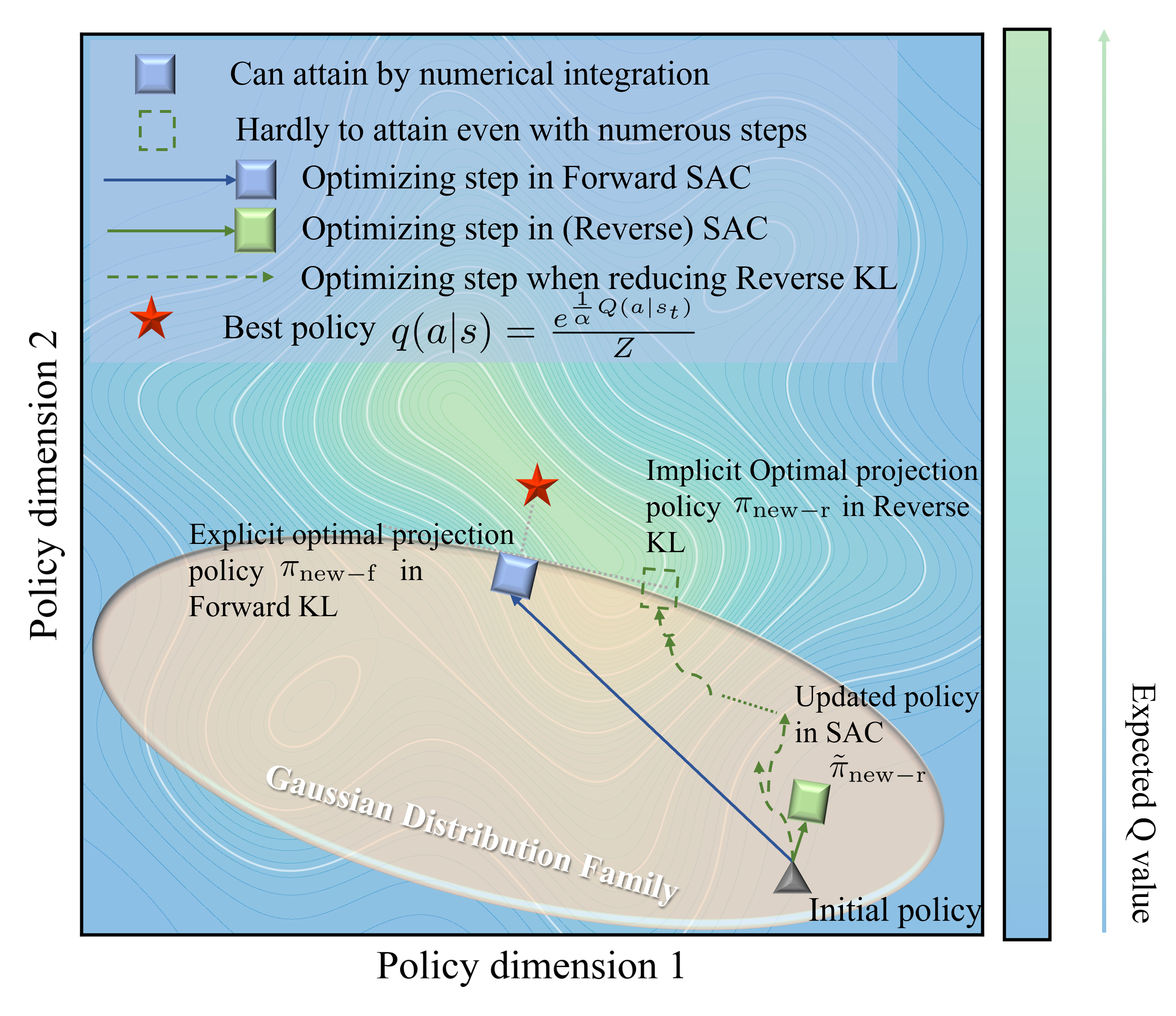} 
 \caption{  }
\label{Abstract1} 
\end{subfigure}
\hfill
\begin{subfigure}[b]{0.49\linewidth}
\centering
\includegraphics[width=\columnwidth]{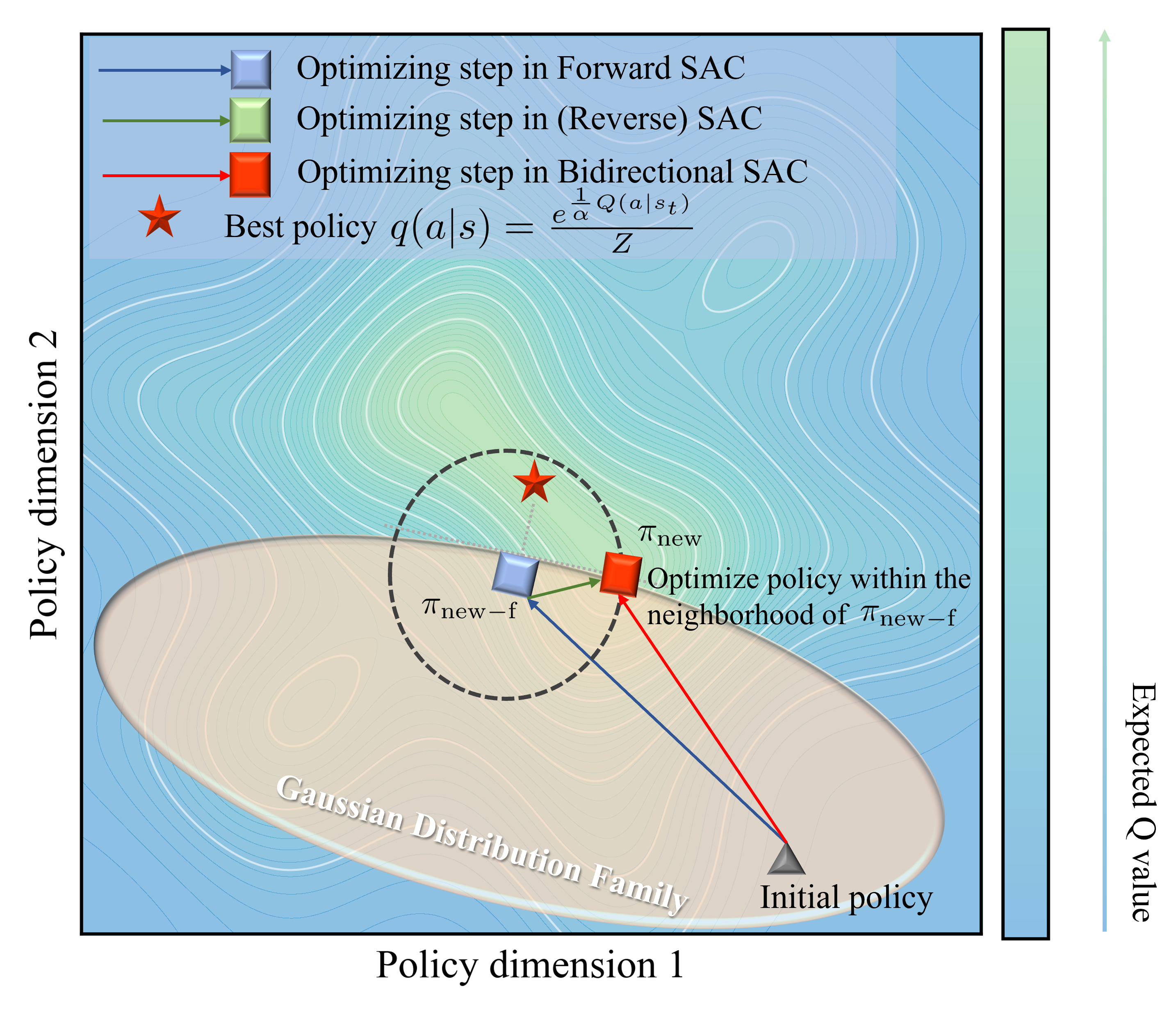} 
\caption{  }
\label{Abstract2} 
\end{subfigure}
\caption{Conceptual illustration of policy optimization dynamics for (Reverse) SAC and Forward SAC. The target Boltzmann distribution is $q(\cdot \mid s_t)$, and the shaded region represents the Gaussian distribution family for policy $\pi^{\prime}$. (a) Comparison of two directions of KL divergence. (b) The schematic diagram of Bidirectional SAC.}
\vspace{-5pt}
\end{figure*}

In this section, we contrast the properties of policy updates in the standard SAC algorithm, which minimizes the reverse KL divergence $D_{\mathrm{KL}}(\pi' \| q)$, with our proposed Forward SAC approach, which minimizes the forward KL divergence $D_{\mathrm{KL}}(q \| \pi')$. 

\paragraph{Forward SAC}
This approach optimizes the policy by minimizing the forward KL divergence $D_{\mathrm{KL}}(q(\cdot \mid s_t) \| \pi'(\cdot \mid s_t))$, as formulated in Equation~\eqref{Projection_problem_inv}. As derived in Section \ref{Analysis}, the optimal projection policy $\pi_{\mathrm{new-f}}$ within the family $\Pi$ that minimizes this divergence can be computed explicitly. Specifically, the optimal mean $f^i(s_t)^*$ and variance $\Sigma^i{}^*$ for each action dimension $i$ correspond to the mean and variance of the marginal target distribution $q_i(\cdot \mid s_t)$, calculable via numerical integration. Thus, the policy $\pi_{\mathrm{new-f}}$ representing the closest projection of $q(\cdot \mid s_t)$ onto $\Pi$ in the forward KL sense is directly attainable in each iteration. However, minimizing $D_{\mathrm{KL}}(q \| \pi')$ does not theoretically guarantee that the resulting policy $\pi_{\mathrm{new-f}}$ maximizes the expected Q value $\mathbb{E}_{a \sim \pi_{\mathrm{new-f}}} [Q^{\pi_{\mathrm{new-f}}}(s_t, a)]$ within the policy family $\Pi$.

\paragraph{(Reverse) SAC}
The standard SAC algorithm performs policy improvement by minimizing the reverse KL divergence $D_{\mathrm{KL}}(\pi'(\cdot \mid s_t) \| q(\cdot \mid s_t))$, as defined in Equation~\eqref{Projection_problem}. Due to the intractability of directly solving this minimization, SAC employs stochastic gradient descent on the policy parameters $\phi$, using the update rule derived from $J_\pi(\phi)$ (approximated in Equation~\eqref{SAC_updating}). As established by \citet{SAC_arxiv}, successfully reducing the reverse KL divergence $D_{\mathrm{KL}}(\pi' \| q)$ guarantees an improvement in the policy objective, ensuring that $\mathbb{E}_{a \sim \pi_{\mathrm{new-r}}} [Q^{\pi_{\text{new-r}}}(s_t, a_t)] \ge \mathbb{E}_{a \sim \pi_{\mathrm{old}}}[Q^{\pi_{\text{old}}}(s_t, a_t)]$. This provides a theoretical basis for monotonic policy improvement. However, the practical gradient-based optimization faces significant challenges. As discussed in Section \ref{Analysis} and illustrated in Figure \ref{visualize_gradient}, estimating the gradient $\nabla_\phi J_\pi(\phi)$ requires sampling actions $a_t$ from the current policy $\pi_\phi(\cdot \mid s_t)$. This sampling process often introduces high variance into the gradient estimates, leading to optimization instability and potentially slow or unreliable convergence towards the true projection $\pi_{\text{new-r}}$ \citep{F_R_KL, F_R_KL_Els}.
Figure~\ref{Abstract1} provides a conceptual visualization of these distinct optimization dynamics within the policy space $\Pi$. The optimization target is the Boltzmann distribution $q(\cdot \mid s_t)$. Both algorithms seek an optimal projection policy within $\Pi$. Forward SAC identifies the policy $\pi_{\mathrm{new-f}}$ that is the direct projection of $q$ onto $\Pi$ with respect to the forward KL divergence. This point is efficiently computed via integration (as shown in our work) but might not coincide with the policy in $\Pi$ that yields the highest expected Q value. Standard SAC, conversely, takes one updating step based on gradients derived from the reverse KL objective (Equation~\eqref{SAC_updating}). While a successful gradient step aims to improve the expected $Q^{\pi_{\text{old}}}$, the practical realization involves noisy updates. 

\subsection{Bidirectional SAC}
It is observed that Forward SAC can find the optimal projection policy $\pi_\text{new-f}$ through forward KL divergence, and SAC can perform policy improvement through minimizing reverse KL divergence. Therefore, in every iteration, by setting the initial policy as the optimal projection policy $\pi_{\mathrm{new-f}}$ in Forward SAC and improving it by optimizing the reverse KL divergence within a certain neighborhood of $\pi_{\mathrm{new-f}}$, Bidirectional SAC is proposed. The optimization problem of Bidirectional SAC is derived as:
\begin{equation}
\label{Bid_update}
     \mathcal{L}(\pi_{})= D_{\mathrm{KL}}\left(\pi_{}\left(\cdot \mid s_t\right)  \bigg\|  q(\cdot \mid s_t) \right) + \epsilon\left[\|f^i(s_t)^* - f_{}^i(s_t)\|^2 + \| \Sigma^i{}^* - \Sigma_{}^i{}\|^2 \right].
\end{equation}
Here, $f^i(s_t)$ and $\Sigma^i$ denote the mean and variance of the policy $\pi$. The optimal projection policy $\pi_{\mathrm{new-f}}$ serves as an initial foundation for the subsequent policy improvement phase. This phase is methodically facilitated through the optimization of reverse KL divergence, as illustrated in Figure~\ref{Abstract2}. The figure demonstrates how Bidirectional SAC explicitly computes the optimal projection policy, which then functions as a strategic starting point. The resultant updated policy $\pi_{\mathrm{new}}$ is optimized within a neighborhood of this initial policy, with the neighborhood's extent regulated by the hyperparameter $\epsilon$. Through this methodological approach, Bidirectional SAC effectively synthesizes the advantages of optimization in both forward and reverse KL divergence directions.

The optimization process within each step initiates with an initial policy that represents the optimal projection onto the Boltzmann distribution $q(\cdot|s_t)$. Subsequently, the algorithm identifies an enhanced policy through the policy improvement step. To efficiently optimize this objective, we employ an actor network $\pi_{\phi}$ trained via the loss function $\mathcal{L}(\pi_{\phi})$. Gradient descent remains our optimization method of choice, implemented by modeling the policy's mean $f_{\phi}^i(s_t)$ and variance $\Sigma_{\phi}^i$ using neural networks. This updating procedure demonstrates enhanced sample efficiency and stability, primarily attributable to the utilization of the MSE loss function. The current implementation establishes a robust foundation while opening avenues for future refinements that would more comprehensively leverage the theoretical advantages of the optimal projection policy. This represents a promising direction for extending the framework, as elaborated in Section \ref{conclusion}. The complete Bidirectional SAC algorithm is formally presented in Algorithm \ref{Bidirectional SAC}.

\begin{algorithm}
\caption{Bidirectional SAC algorithm}
\label{Bidirectional SAC}
\textbf{Hyperparameters:} Total number of training steps $\mathcal{L}$, mini-batch size $\mathcal{M}$, the update epoch $\mathcal{J}$ of neural networks  each training, and the learning rate $\beta$
\begin{algorithmic}[1]  
    \STATE Initialize the critic network  $Q_{\theta}$, actor network $\pi_{\phi}$ and replay buffer $\mathcal{B}$.
    \STATE Start with the initial state $s_0$
    \FOR{$l = 1, \dots ,\mathcal{L}$}
        \STATE Using policy $\pi_{\phi}$, collect and store transitions $(s_t, a_t, r_t, s_{t+1})$ in replay buffer $\mathcal{B}$.
        \FOR{$j=1,\dots,\mathcal{J}$}
        \STATE Sample mini-batch $\{(s_i, a^i, r_i, s_{i+1}) \mid i = 1, \dots, \mathcal{M}\}$ from $\mathcal{B}$.
        \STATE Update the critic network $Q_{\theta}$ by $\theta \leftarrow \theta-\beta \nabla_\theta J_Q(\theta)$ in \eqref{Q_updated}
        \STATE Update the actor parameters with $\phi \leftarrow \phi-\beta \nabla_{\phi} \mathcal{L}(\pi_{\phi})$ in \eqref{Bid_update}. 
        \ENDFOR
    \ENDFOR
    
\end{algorithmic}
\end{algorithm}




\section{Experiment result}
\label{Exp}

This section evaluates the episodic reward of Bidirectional SAC, including an ablation study against traditional SAC and Forward SAC. We benchmark performance on several MuJoCo environments \citep{MuJoCo} (Swimmer, Hopper, Walker2D, Ant, Humanoid, and HumanoidStandup), chosen for their varying complexity and common reinforcement learning objectives such as locomotion and stability \citep{ICLR_PPO}. Additional results from other MuJoCo and Box2D \citep{Box2D} tasks are in Appendix \ref{Effectiveness}.

\subsection{Overall performance}
\label{overall_performance}
\begin{figure}[htbp]
    \centering
    \includegraphics[width=\linewidth]{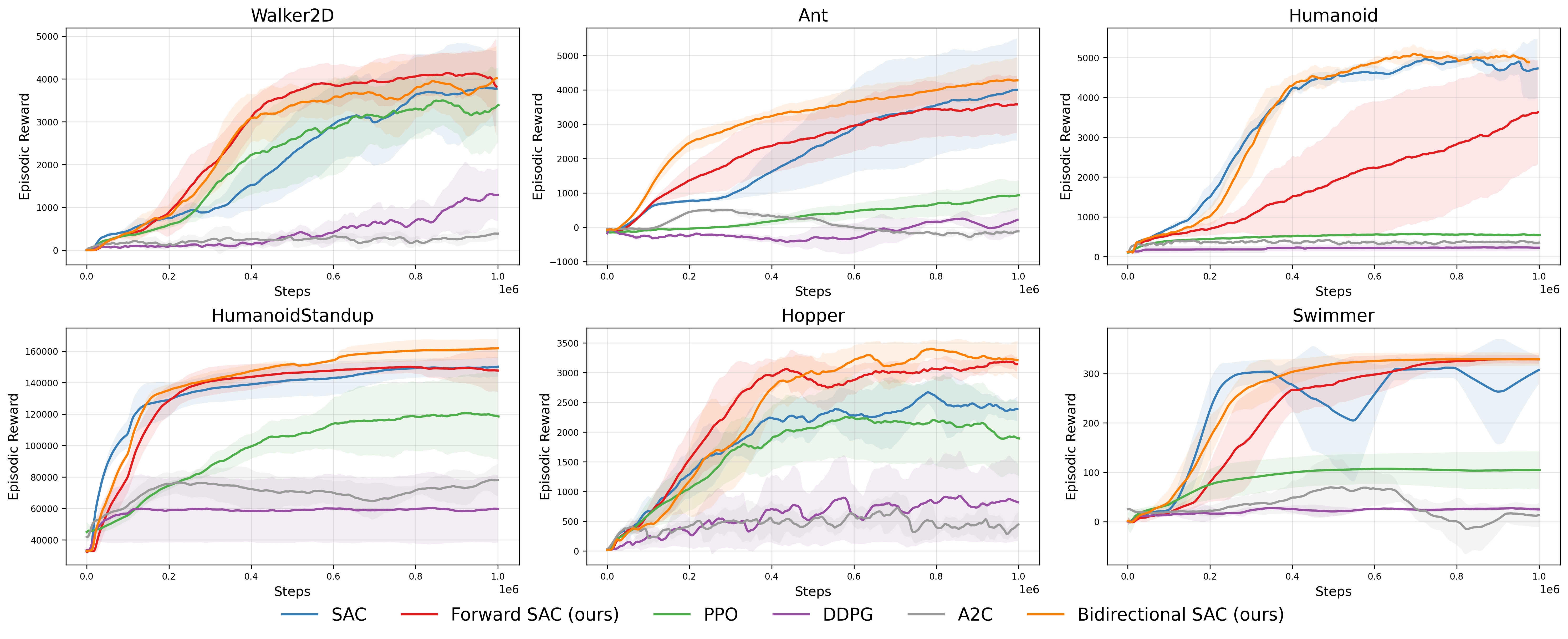}
    \caption{The episodic reward in MuJoCo environments for Bidirectional SAC, Forward SAC, and other benchmarks in the training process.}
    \label{Compare_all}
\end{figure}
We compare the performance of the Bidirectional SAC algorithm with the traditional SAC algorithm and Forward SAC algorithm, as well as established benchmarks, including A2C \citep{A2C}, PPO \citep{ICLR_PPO}, and DDPG \citep{DDPG}. The results are shown in Figure \ref{Compare_all}. It is demonstrated in the figure that both Bidirectional SAC and Forward SAC enjoy faster convergence compared to traditional SAC, which is deemed as sample efficient because it converges faster than other benchmarks. Compared to the benchmark algorithms, Forward SAC attains better overall performance in simple and medium complex environments, including Swimmer, Hopper, and Walker2D. However, it can only perform mediocre or even worse in other complex environments. This may be due to that in complex environments, the distribution landscape of Boltzmann distribution $q(\cdot|s_t)$ is too complex that even the optimal policy with respect to Forward KL divergence could cause a substantial difference in the performance of the policy. This issue can be overcome by the Bidirectional SAC algorithm, which searches for better performance policies with the optimal policy initialization through integrating both directions of KL divergence.
\subsection{Effectiveness analysis}
\label{Effectiveness}
\begin{figure}[htbp]
    \centering
    \includegraphics[width=\linewidth]{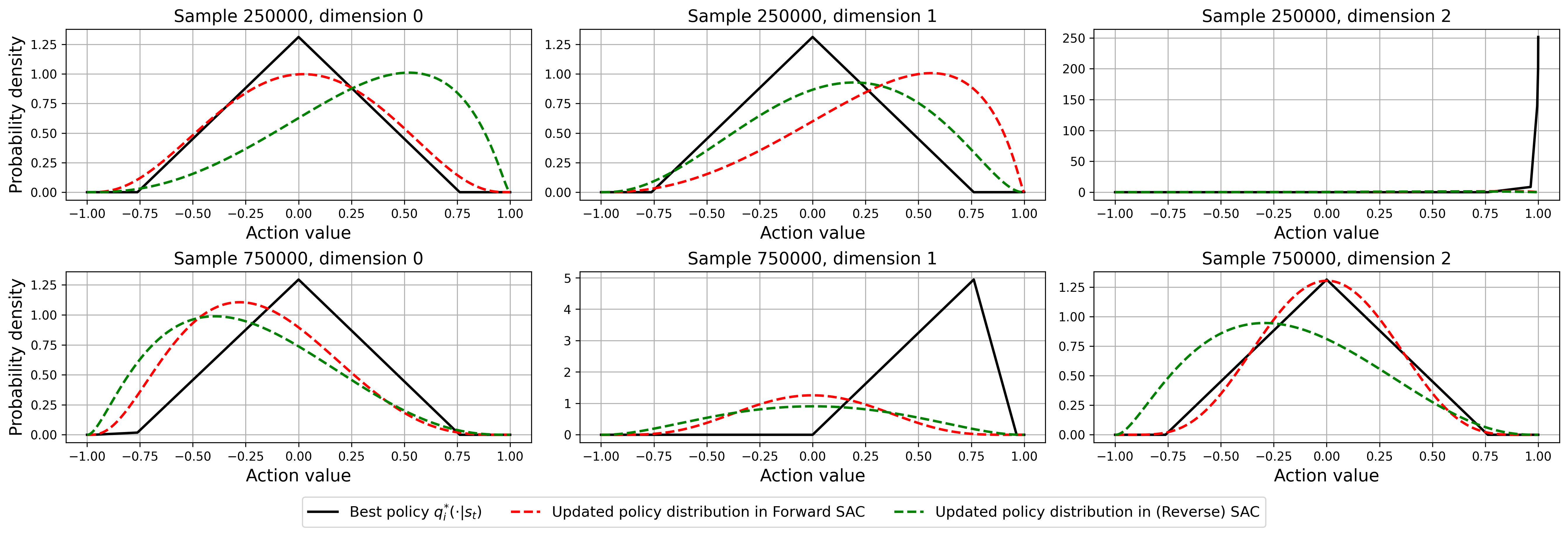}
    \caption{The comparison result for the updated policy in Forward SAC and standard SAC to the Boltzmann distribution $q(a_t|s_t)$.}
    \label{integrated performance compare}
    \vspace{-15pt}
\end{figure}

To rigorously analyze the effectiveness of the proposed Bidirectional SAC algorithm, and specifically the Forward SAC component, two key questions must be addressed: (1) How accurately does the VDN-a network learn the marginal distribution? (2) Is the optimal projection policy $\pi_{\mathrm{new-f}}$, derived by optimizing the forward KL divergence, demonstrably closer to the target Boltzmann distribution $q(\cdot |s_t)$ than the traditional updated policy $\pi_{\mathrm{new-r}}$, obtained via gradient optimization of reverse KL divergence?
\begin{figure}[htbp]
    \centering
    \includegraphics[width=\linewidth]{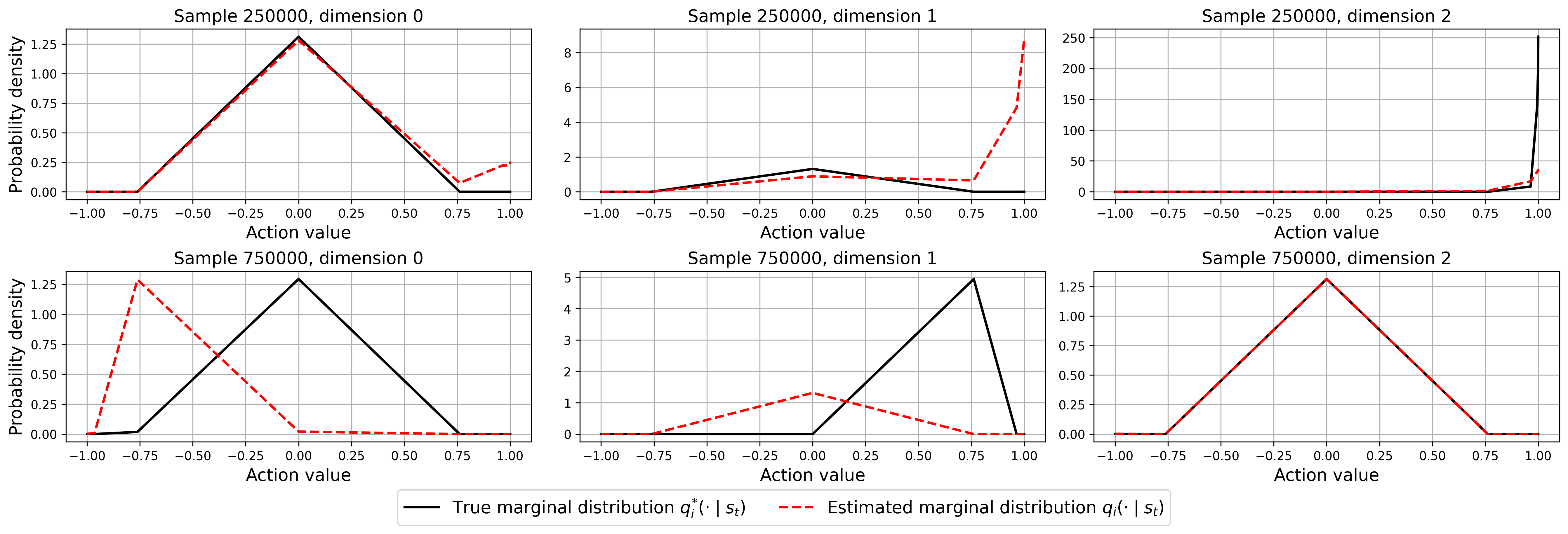}
    \caption{The comparison result in marginal distribution learned by VDN-a networks to the true distribution.}
    \label{VDN-a estiamted}
    \vspace{-15pt}
\end{figure}
To provide insight into these questions, we examine the training process of Bidirectional SAC within the Hopper environment. The Hopper environment's action space is three-dimensional, which facilitates analysis and serves as a suitable testbed for validating our claims. We analyze data from two distinct updating steps during training. Figure \ref{VDN-a estiamted} presents a comparison between the estimated marginal distribution $q_i(\cdot \mid s_t)$ and the ground truth marginal distribution $q_i^*(\cdot \mid s_t)$. The true marginal distribution is readily computed via multidimensional numerical integration of $q(\cdot \mid s_t)$. As depicted in the figure, while the estimated marginal distribution $q_i(\cdot \mid s_t)$ exhibits some deviation from the true distribution, it successfully captures the principal characteristics of the ground truth, particularly the mean and variance information critical for the updating mechanism in Bidirectional SAC. Addressing the second question, Figure \ref{integrated performance compare} illustrates a comparison of the updated policy distributions from Forward SAC and standard SAC at the same training step. The figure clearly indicates that the updated policy distribution resulting from Forward SAC more closely resembles the target distribution $q(\cdot |s_t)$ than the policy updated using the standard SAC algorithm's reverse KL approach, thereby answering the second question affirmatively. Further comparisons of policy distributions during the updating steps are provided in Appendix \ref{dis_comparison}.

\section{Conclusion and future work}
\label{conclusion}
This paper addressed the optimization challenges of reverse KL divergence in the SAC algorithm. We demonstrated that employing forward KL divergence permits an explicit optimal policy projection, calculable via numerical integration and facilitated by our proposed VDN-a critic architecture. Building upon this insight, we introduced Bidirectional SAC, an algorithm that combines the direct policy computation inherent in forward KL with the policy improvement guarantees associated with reverse KL. Empirical results show that Bidirectional SAC significantly surpasses standard SAC and other baselines in both sample efficiency and asymptotic performance, achieving up to a $30\%$ increase in episodic rewards.

Despite these advancements, we acknowledge that the optimal projection policy in the forward KL component is approximated using the VDN-a network, and the actor updates in Bidirectional SAC still rely on gradient descent due to neural network parameterization. While our extensive experiments validate the superior performance of Bidirectional SAC, further research could aim to bridge the gap between these practical approximations and the theoretically true optimal projection policy. Exploring methods for even more accurate policy projections and refined optimization techniques within this bidirectional framework presents a promising avenue for future investigation.

\bibliographystyle{plainnat}  
\bibliography{example_paper}


\appendix

\section{KL divergence derivation}
\label{KL_partial_diri}
We need to compute the partial derivatives of the objective function
$$
D_{\mathrm{KL}}\left( q(\cdot \mid s_t) \| \pi^{\prime}(a \mid s_t) \right)
$$
with respect to $ f_\phi^i(s_t) $ and $ \Sigma_\phi^i $. Given that $ \pi^{\prime}(a \mid s_t) = \prod_i \pi_i^{\prime}(a^i \mid s_t) $, where each $ \pi_i^{\prime}(a^i \mid s_t) = \mathcal{N}(a^i; f_\phi^i(s_t), \Sigma_\phi^i) $, we will derive the necessary expressions by considering the marginal normal distributions' mean and variance.

\subsection{Expanding the KL divergence expression}
According to the definition of KL divergence:
$$
D_{\mathrm{KL}}\left( q(\cdot \mid s_t) \| \pi^{\prime}(a \mid s_t) \right) = \int_{\mathcal{A}} q(a \mid s_t) \log q(a \mid s_t) - q(a \mid s_t) \log \pi^{\prime}(a \mid s_t) \, da.
$$
Since $ \pi^{\prime}(a \mid s_t) $ is a product of individual components:
$$
\pi^{\prime}(a \mid s_t) = \prod_i \pi_i^{\prime}(a^i \mid s_t),
$$
We have:
$$
\log \pi^{\prime}(a \mid s_t) = \sum_i \log \pi_i^{\prime}(a^i \mid s_t).
$$
Thus, the KL divergence becomes:
\begin{equation}
\begin{aligned}
       &D_{\mathrm{KL}}\left( q(\cdot \mid s_t) \| \pi^{\prime}(a \mid s_t) \right) = \int_{\mathcal{A}} q(a \mid s_t) \log q(a \mid s_t) \, da \\
       &+ \sum_i \frac{1}{2} \log(2 \pi \Sigma_\phi^i) \int_{\mathcal{A}} q(a \mid s_t) \, da - \sum_i \frac{1}{2 \Sigma_\phi^i} \int_{\mathcal{A}} q_i(a^i \mid s_t) (a^i - f_\phi^i(s_t))^2 \, da^i. 
\end{aligned}
\end{equation}

Since $ \int_{\mathcal{A}} q(a \mid s_t) da = 1 $, the second term simplifies to:
$$
\sum_i \frac{1}{2} \log(2 \pi \Sigma_\phi^i).
$$
The third term corresponds to the variance term of $ q_i(a^i \mid s_t) $.

\subsection{Partial derivatives with respect to $ f_\phi^i(s_t) $ and $ \Sigma_\phi^i $}

\subsubsection{Derivative with respect to $ f_\phi^i(s_t) $}

We need to compute the terms involving $ f_\phi^i(s_t) $. The relevant term is:
$$
\sum_i \frac{1}{2 \Sigma_\phi^i} \int_{\mathcal{A}} q_i(a^i \mid s_t) (a^i - f_\phi^i(s_t))^2 \, da^i.
$$
Taking the derivative with respect to $ f_\phi^i(s_t) $, we get:
$$
\frac{\partial}{\partial f_\phi^i(s_t)} \left( \frac{1}{2 \Sigma_\phi^i} \int_{\mathcal{A}_i} q_i(a^i \mid s_t) (a^i - f_\phi^i(s_t))^2 \, da^i \right) = \frac{1}{\Sigma_\phi^i} \int_{\mathcal{A}_i} q_i(a^i \mid s_t) (a^i - f_\phi^i(s_t)) \, da^i.
$$
This is the derivative of the mean $ f_\phi^i(s_t) $ of the marginal normal distribution, which corresponds to the first moment (i.e., the mean) of $ q_i(a^i \mid s_t) $.

\subsubsection{Derivative with respect to $ \Sigma_\phi^i $}

To compute the derivative with respect to $ \Sigma_\phi^i $, we differentiate each term:
$$
\frac{\partial}{\partial \Sigma_\phi^i} \left( \frac{1}{2} \log(2 \pi \Sigma_\phi^i) \right) = \frac{1}{2 \Sigma_\phi^i}.
$$
For the second term:
$$
\frac{\partial}{\partial \Sigma_\phi^i} \left( \frac{1}{2 \Sigma_\phi^i} \int_{\mathcal{A}_i} q_i(a^i \mid s_t) (a^i - f_\phi^i(s_t))^2 \, da^i \right) = -\frac{1}{2 (\Sigma_\phi^i)^2} \int_{\mathcal{A}_i} q_i(a^i \mid s_t) (a^i - f_\phi^i(s_t))^2 \, da^i.
$$
Thus, the partial derivative with respect to $ \Sigma_\phi^i $ is:
$$
\frac{\partial D_{\mathrm{KL}}}{\partial \Sigma_\phi^i} = \frac{1}{2 \Sigma_\phi^i} - \frac{1}{2 (\Sigma_\phi^i)^2} \int_{\mathcal{A}_i} q_i(a^i \mid s_t) (a^i - f_\phi^i(s_t))^2 \, da^i.
$$

\section{Squashing function}
\label{tanh_s}
In case the action space is typically bounded by using an invertible squashing function, the optimal policy projection calculation should be adjusted for this transformation.

Recalling the definition of KL divergence:
$$
D_{\mathrm{KL}}\left( q(\cdot \mid s_t) \| \pi^{\prime}(a \mid s_t) \right) = \int_{\mathcal{A}} q(a \mid s_t) \log q(a \mid s_t) - q(a \mid s_t) \log \pi^{\prime}(a \mid s_t) \, da.
$$
If we use a transformation $a = h(x)$, where $h(\cdot)$ is the squashing function to transform the original action value $x$ to fit the action space in a specific environment. Without loss of generality, we assume $h(\cdot)$ is consistent in each dimension of the action, which means that $(a^1,a^2,\dots,a^N) = (h(x^1),h(x^2),\dots,h(x^N))$.  Therefore, the probability density function is given by:
$$
\pi^{\prime}(a^i \mid s_t) = \pi^{\prime}(x^i \mid s_t) \left| \frac{dx^i}{dh(x^i)} \right| = \frac{1}{\sqrt{2\pi\Sigma_\phi^i}} \exp\left( -\frac{(x^i - f_\phi^i(s_t))^2}{2\Sigma_\phi^i} \right) \left| \frac{1}{h'(x^i)} \right|
$$

Substituting  $\pi(y)$ into the KL divergence, we can obtain the partial derivative with respect to $\mu$ as:
\begin{equation}
    \begin{aligned}
   \frac{\partial D_{\mathrm{KL}}\left( q(\cdot \mid s_t) \| \pi^{\prime}(a \mid s_t) \right)}{\partial f_\phi^i(s_t)} &= -\int_{\mathcal{A}} q(a \mid s_t) \frac{\partial}{\partial f_\phi^i(s_t)} \log(\pi^{\prime}(h^{-1}(a) \mid s_t)) da \\
   & = -\int_{-\infty}^{\infty} q(h(x^i) \mid s_t) \frac{\partial}{\partial f_\phi^i(s_t)} \log(\pi^{\prime}(x^i \mid s_t)) h^{'}(x^i) dx^i \\
   & = \frac{1}{\Sigma_\phi^i} \int_{\mathcal{A}} q(h(x^i) \mid s_t) (x^i - f_\phi^i(s_t)) h^{'}(x^i) \, dx^i
\end{aligned}
\nonumber
\end{equation}

Therefore, it is easily derived that the optimal mean value $f_\phi^i(s_t)$ is
$$
f_\phi^i(s_t) = \int_{-\infty}^{\infty} q_i(h(x^i) \mid s_t) \cdot x^i \cdot h^{'}(x^i)\, dx^i.
$$
Similarly, the partial derivative with respect to $ \Sigma_\phi^i $ is:
\begin{equation}
\frac{\partial D_{\mathrm{KL}}\left( q(\cdot \mid s_t) \| \pi^{\prime}(a \mid s_t) \right)}{\partial \Sigma_\phi^i} = \frac{1}{2 \Sigma_\phi^i} - \frac{1}{2 (\Sigma_\phi^i)^2} \int_{-\infty}^{\infty} q_i(h(x^i) \mid s_t) (x^i - f_\phi^i(s_t))^2 \, dx^i .
\nonumber
\end{equation}
Therefore, the optimal variance 
\begin{equation}
  \begin{aligned}
    \Sigma_\phi^i &= \int_{-\infty}^{\infty} q_i(h(x^i) \mid s_t) (x^i - f_\phi^i(s_t))^2 \, dx^i 
    \nonumber
\end{aligned}  
\end{equation}

By substituting the squashing function $h(\cdot)$ with $\tanh(\cdot)$, the ultimate optimal projection policy in Equation~\eqref{final_tanh} is easily attained.

\section{The VDN-a network}
\subsection{The introduction to the VDN network}
\label{VDN_net}
To compute the optimal policy parameters $f^{i}(s_t)^{*}$ and $\Sigma^{i}{}^{*}$ (mean and variance of the target Boltzmann distribution's marginals), various methods could be considered. Importance sampling \citep{MC_book}, common in policy optimization \citep{IS_1} and regularized reinforcement learning \citep{UREX, ECPO}, faces challenges in identifying a suitable proposal distribution that closely approximates the target \citep{UREX}. Numerical integration offers an alternative, but standard multidimensional integration suffers from the curse of dimensionality \citep{Int1}, making it computationally prohibitive for high-dimensional action spaces. Even if $f^{i}(s_t)^{*}$ and $\Sigma^{i}{}^{*}$ are computed per dimension, obtaining the true marginal distribution $q_i(\cdot \mid s_t)$ of the full Boltzmann distribution $q(\cdot \mid s_t)$ remains difficult due to the complex dependencies between action dimensions.

\begin{figure}[htbp]
    \centering
    \includegraphics[width=0.5\linewidth]{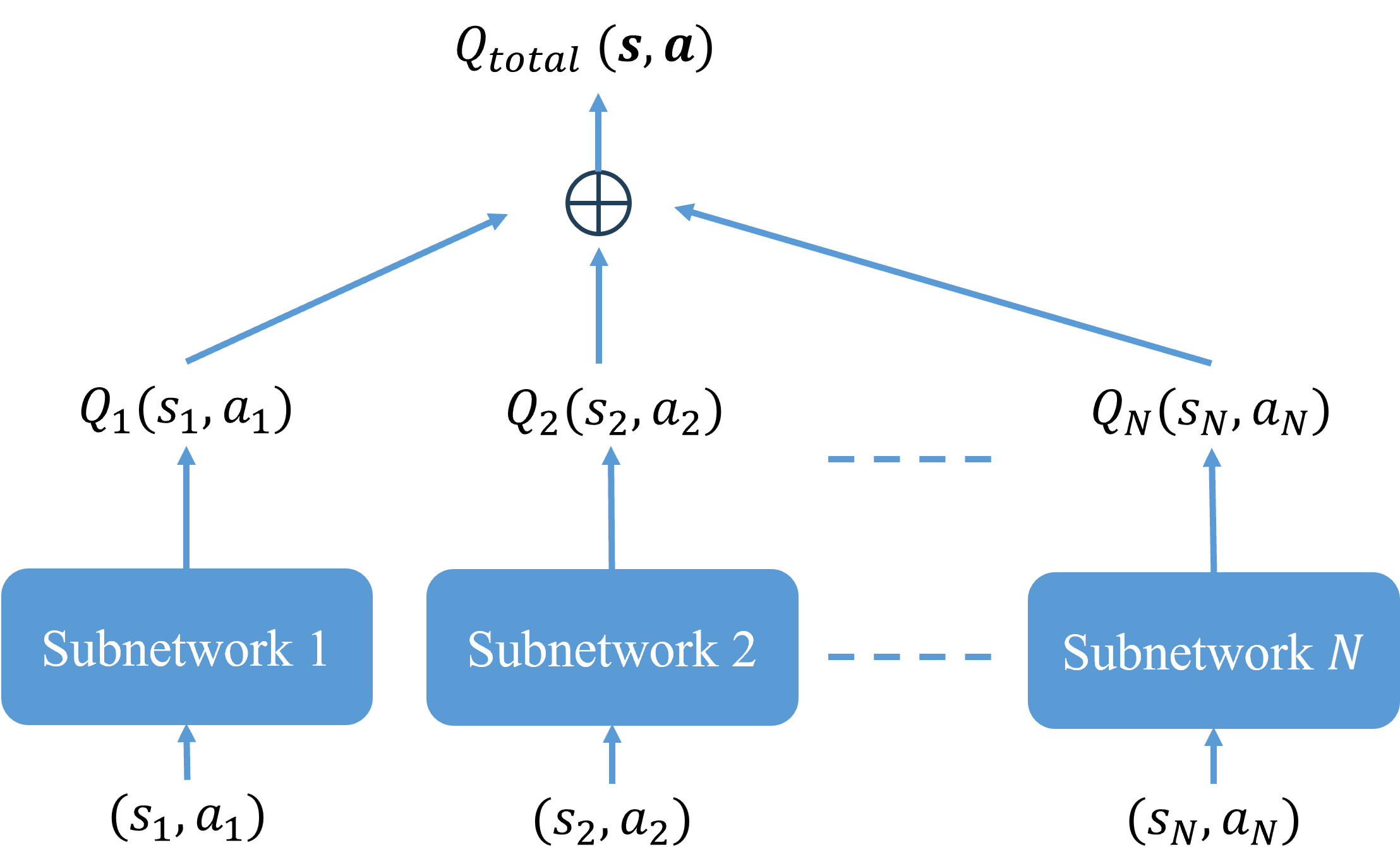} 
    \caption{Framework of the original Value Decomposition Network (VDN) for MARL.}
    \label{VDN}
\end{figure}

To address these challenges, we propose the VDN-a network. This architecture is inspired by Value Decomposition Networks (VDN) \citep{VDN}, a significant advancement in Multi-Agent Reinforcement Learning (MARL) for cooperative systems with decentralized execution. The core idea of VDN, illustrated in Figure~\ref{VDN}, is to decompose a global value function $Q_{total}(s, a)$ into a sum of individual agent-specific value functions $Q_i(s_i, a_i)$:
\begin{equation}
    Q_{total}(s, a) = \sum_{i=1}^N Q_i(s_i, a_i)
\end{equation}
where $s_i$ and $a_i$ are the local observation and action for agent $i$. This additivity allows each agent to make decisions based on its local value function, while the system is trained centrally. VDNs support scalability and are effective in cooperative scenarios where the team's objective can be approximated as a sum of individual contributions, offering benefits like reduced complexity and improved credit assignment.

The additive structure of VDN has proven effective in MARL and provides a foundation for more sophisticated value decomposition methods. In our single-agent context, we adapt this principle in the VDN-a network (detailed in Section 3.3 of the main paper and Figure \ref{fig:VDN-a network}) not to decompose by agent, but to structure the Q-function $Q(s_t, a_t)$ such that components primarily related to individual action dimensions $a_t^i$ can be learned by subnetworks $Q_i(s_t, a_t^i)$, with an auxiliary network $U(s_t, a_t)$ capturing interactions. This facilitates the estimation of terms related to the marginals of the Boltzmann distribution, as theoretically justified below.

\subsection{Theoretical foundation for VDN-a network using copula theory}
\label{theoretic_copula}

This section provides a theoretical framework, grounded in Copula theory, to explain how the VDN-a network architecture facilitates the learning of components essential for recovering the true marginal distributions of the Boltzmann target distribution $q(a|s_t)$. This capability is crucial for the Forward SAC component of our Bidirectional SAC algorithm, which relies on the mean and variance of these marginals.

\subsubsection{Sklar's theorem and preliminary definitions}
Let $a = (a_1, a_2, \dots, a_n)$ be an $n$-dimensional action vector in $\mathcal{A} \subseteq \mathbb{R}^n$ for a given state $s_t \in \mathcal{S}$.
\begin{itemize}
    \item $Q^{\pi_{\text{old}}}(s_t, a)$: The soft Q-function from the previous policy $\pi_{\text{old}}$.
    \item $Q_{\text{Boltz}}(a; s_t) \equiv \frac{1}{\alpha} Q^{\pi_{\text{old}}}(s_t, a)$: The potential function for the Boltzmann distribution at state $s_t$, where $\alpha$ is the temperature parameter. For conciseness, $Q_{\text{Boltz}}(a)$ is used when $s_t$ is implicit.
    \item $Z(s_t) = \int_{\mathcal{A}} \exp(Q_{\text{Boltz}}(a'; s_t)) da'$: The partition function for $q(a|s_t)$.
    \item $q(a|s_t) = \frac{\exp(Q_{\text{Boltz}}(a; s_t))}{Z(s_t)}$: The target Boltzmann joint PDF over actions.
    \item $M^{(j)}(a_j; s_t) \equiv \int_{\mathcal{A}_{-j}} \exp(Q_{\text{Boltz}}(a_j, a'_{-j}; s_t)) da'_{-j}$: The marginalization of $\exp(Q_{\text{Boltz}}(a; s_t))$ over action dimensions except $a_j$. The term $\ln(M^{(j)}(a_j; s_t))$ is central to what VDN-a subnetworks aim to capture.
    \item $q_j(a_j|s_t) = \frac{M^{(j)}(a_j; s_t)}{Z(s_t)}$: The $j$-th true marginal PDF of $q(a|s_t)$.
    \item $F_j(a_j|s_t) = \int_{-\infty}^{a_j} q_j(t_j|s_t) dt_j$: The $j$-th true marginal CDF.
    \item $c(u_1, \dots, u_n; s_t)$: The unique copula density function associated with $q(a|s_t)$.
    \item $\mathcal{L}_{c}(a; s_t) \equiv \ln(c(F_1(a_1|s_t), \dots, F_n(a_n|s_t); s_t))$: The log-copula density term.
\end{itemize}
Sklar's Theorem states that any joint PDF $q(a|s_t)$ can be expressed via its marginal CDFs $F_j(a_j|s_t)$ and a unique copula density $c(\cdot ; s_t)$. In terms of densities:
\begin{equation}
q(a|s_t) = c(F_1(a_1|s_t), \dots, F_n(a_n|s_t); s_t) \prod_{j=1}^{n} q_j(a_j|s_t)
\label{eq:sklar_density_form_appendix_revised}
\end{equation}

\subsubsection{Derivation of the structural expression for $Q_{\text{Boltz}}(a; s_t)$}
We decompose $Q_{\text{Boltz}}(a; s_t)$ using Eq. \eqref{eq:sklar_density_form_appendix_revised}.
Given $q(a|s_t) = \exp(Q_{\text{Boltz}}(a; s_t))/Z(s_t)$, equating this with Eq. \eqref{eq:sklar_density_form_appendix_revised} and taking logarithms yields:
$$ Q_{\text{Boltz}}(a; s_t) - \ln(Z(s_t)) = \ln(c(F_1(a_1|s_t), \dots, F_n(a_n|s_t); s_t)) + \sum_{j=1}^{n} \ln(q_j(a_j|s_t)) $$
Using $\mathcal{L}_{c}(a; s_t) = \ln(c(\dots))$ and substituting $q_j(a_j|s_t) = M^{(j)}(a_j; s_t)/Z(s_t)$:
$$ Q_{\text{Boltz}}(a; s_t) = \ln(Z(s_t)) + \sum_{j=1}^{n} \left(\ln(M^{(j)}(a_j; s_t)) - \ln(Z(s_t))\right) + \mathcal{L}_{c}(a; s_t) $$
This simplifies to:
\begin{equation}
Q_{\text{Boltz}}(a; s_t) = (1-n)\ln(Z(s_t)) + \sum_{j=1}^{n} \ln(M^{(j)}(a_j; s_t)) + \mathcal{L}_{c}(a; s_t)
\label{eq:Q_boltz_decomp_appendix_revised}
\end{equation}
This exact decomposition reveals $Q_{\text{Boltz}}(a; s_t)$ as a sum of a state-dependent constant, terms $\ln(M^{(j)}(a_j; s_t))$ (each depending only on $a_j$ and $s_t$), and an interaction term $\mathcal{L}_{c}(a; s_t)$ capturing multivariate dependencies.

\subsubsection{VDN-a network and its connection to the Copula decomposition}
The VDN-a network approximates $Q^{\pi_{\text{old}}}(s_t, a)$ as $Q^{\text{VDN-a}}(s_t, a) = \sum_{j=1}^{n} Q_j^{\text{sub}}(s_t, a_j) + U^{\text{aux}}(s_t, a)$.
The corresponding learned potential function is $Q_{\text{Boltz}}^{\text{VDN-a}}(a; s_t) = \frac{1}{\alpha}Q^{\text{VDN-a}}(s_t, a)$, so:
$$ Q_{\text{Boltz}}^{\text{VDN-a}}(a; s_t) = \sum_{j=1}^{n} \underbrace{\left(\frac{1}{\alpha}Q_j^{\text{sub}}(s_t, a_j)\right)}_{\equiv \hat{Q}_j^*(a_j; s_t)} + \underbrace{\left(\frac{1}{\alpha}U^{\text{aux}}(s_t, a)\right)}_{\equiv \hat{U}^*(a; s_t)} $$
If $Q_{\text{Boltz}}^{\text{VDN-a}}(a; s_t) \approx Q_{\text{Boltz}}(a; s_t)$, the VDN-a structure aims for:
\begin{itemize}
    \item Each $\hat{Q}_j^*(a_j; s_t)$ to primarily learn $\ln(M^{(j)}(a_j; s_t))$ plus absorbable state-dependent constants $k_j(s_t)$. Thus, $\exp(\hat{Q}_j^*(a_j; s_t)) \propto M^{(j)}(a_j; s_t)$. Crucially, $Q_j^{\text{sub}}(s_t, a_j)$ is not an isolated "marginal Q-value" but learns the component of $Q^{\pi_{\text{old}}}$ corresponding to $\ln(M^{(j)})$, given $U^{\text{aux}}$ handles interactions.
    \item $\hat{U}^*(a; s_t)$ to primarily learn the log-copula density $\mathcal{L}_{c}(a; s_t)$ and remaining constants from Eq. \eqref{eq:Q_boltz_decomp_appendix_revised}.
\end{itemize}

\subsubsection{Recovery of true marginal distributions}
The estimated marginal distributions are $\hat{q}_j(a_j|s_t) = \frac{\exp(\hat{Q}_j^*(a_j; s_t))}{\int_{\mathcal{A}_j} \exp(\hat{Q}_j^*(a'_j; s_t)) da'_j}$.
If $\exp(\hat{Q}_j^*(a_j; s_t)) = K_j(s_t) M^{(j)}(a_j; s_t)$ for some $K_j(s_t) > 0$, the denominator becomes $K_j(s_t) \int_{\mathcal{A}_j} M^{(j)}(a'_j; s_t) da'_j$.
Since $\int_{\mathcal{A}_j} M^{(j)}(a'_j; s_t) da'_j = Z(s_t)$, then:
$$ \hat{q}_j(a_j|s_t) = \frac{K_j(s_t) M^{(j)}(a_j; s_t)}{K_j(s_t) Z(s_t)} = \frac{M^{(j)}(a_j; s_t)}{Z(s_t)} = q_j(a_j|s_t) $$
Thus, if each $\frac{1}{\alpha}Q_j^{\text{sub}}(s_t, a_j)$ effectively learns $\ln(M^{(j)}(a_j; s_t))$ (up to an additive constant, implying $\exp(\frac{1}{\alpha}Q_j^{\text{sub}})$ captures $M^{(j)}$ up to a multiplicative constant), and $U^{\text{aux}}$ isolates dependencies, then this procedure recovers the true marginal PDFs.

\begin{figure}[htbp]
    \centering
    \includegraphics[width=0.8\linewidth]{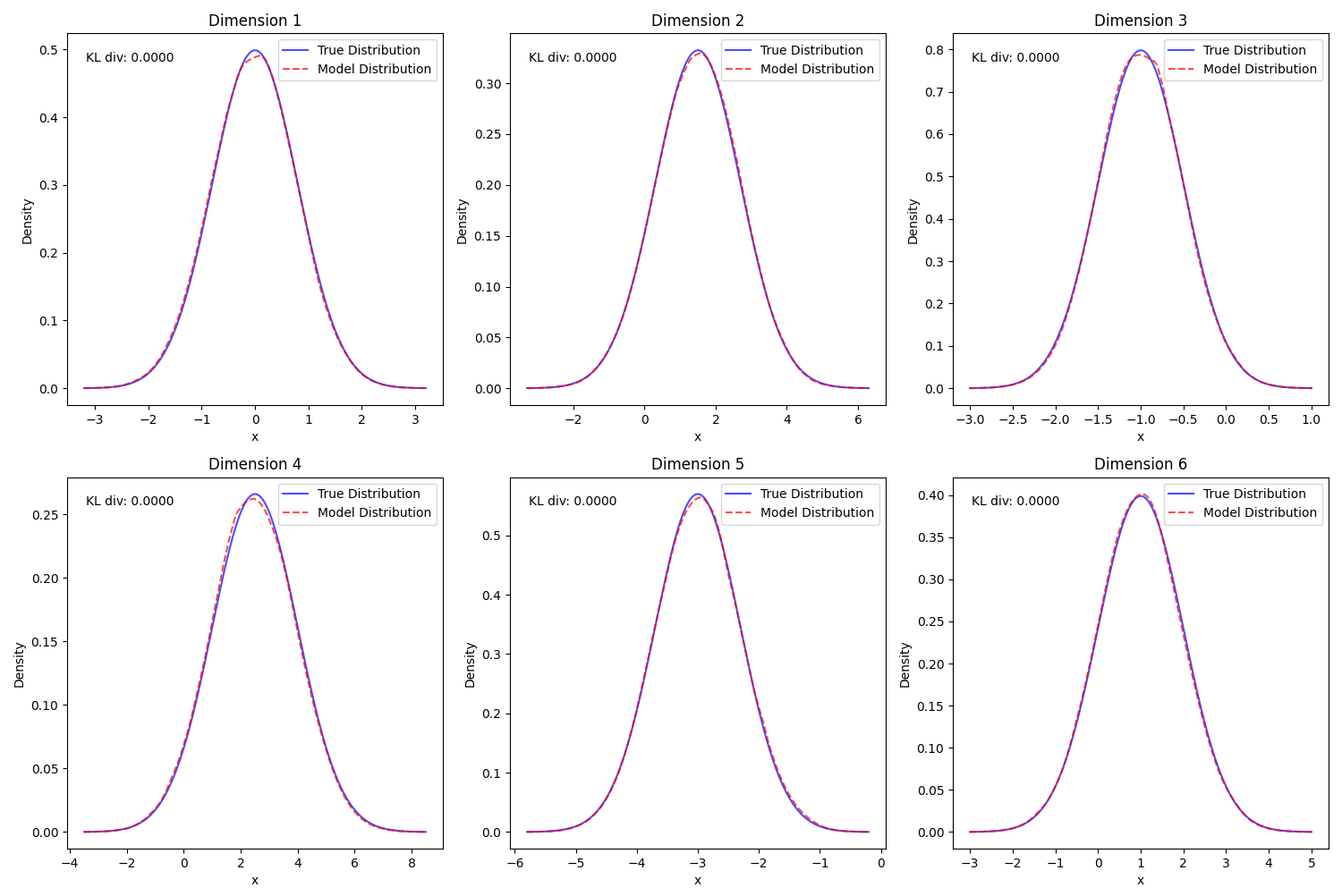}
    \caption{VDN-a performance in approximating components related to a unimodal marginal distribution. (Assumed: Solid lines might be true components/PDF, dashed are VDN-a estimates - clarify in actual caption based on what is plotted).}
    \label{single_modal}
\end{figure}

\begin{figure}[htbp]
    \centering
    \includegraphics[width=0.8\linewidth]{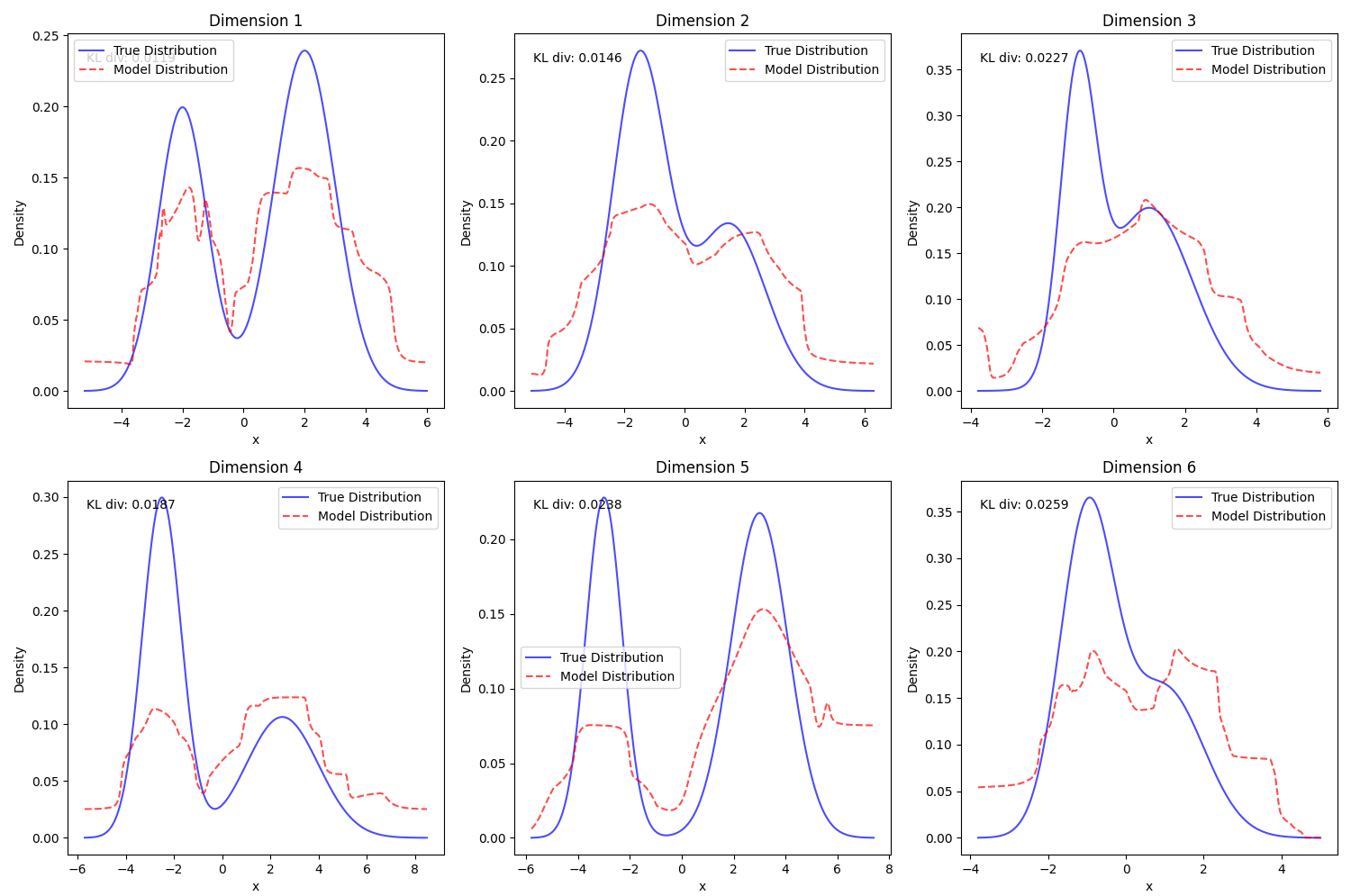}
    \caption{VDN-a performance in approximating components related to a bimodal marginal distribution. (Assumed: Solid lines might be true components/PDF, dashed are VDN-a estimates - clarify in actual caption).}
    \label{bimodal}
\end{figure}

\textbf{Condition of a "sufficiently learned" auxiliary network or independence:}
The main paper's assertion that "if the auxiliary network is sufficiently learned or each dimension of the action is independent, the learned $q_i(a_t^i \mid s_t)$ is actually the true marginal distribution of the $q(\cdot \mid s_t)$" is supported:
\begin{itemize}
    \item \textbf{Sufficiently learned $U^{\text{aux}}$:} If $U^{\text{aux}}$ correctly models $\mathcal{L}_c(a;s_t)$ and other non-separable terms, $Q_j^{\text{sub}}$ can accurately represent components related to $M^{(j)}(a_j;s_t)$, leading to correct $q_j(a_j|s_t)$ recovery.
    \item \textbf{Independence:} If action dimensions are independent, $c(\dots;s_t) = 1$, so $\mathcal{L}_{c}(a;s_t) = 0$. Then $Q_{\text{Boltz}}(a; s_t) = (1-n)\ln(Z(s_t)) + \sum_{j=1}^{n} \ln(M^{(j)}(a_j; s_t))$. If $\hat{U}^*(a;s_t)$ learns the resulting constant term, each $\hat{Q}_j^*(a_j;s_t)$ more directly approximates $\ln(M^{(j)}(a_j;s_t))$, making marginal recovery robust.
\end{itemize}
Figures \ref{single_modal} and \ref{bimodal} illustrate VDN-a's empirical performance in approximating marginal distribution components.
Figure \ref{single_modal} shows effective learning for unimodal target marginal components, while Figure \ref{bimodal} indicates that performance can be challenged by more complex, e.g., bimodal, marginal structures. This challenge may arise from the increased difficulty for $U^{\text{aux}}$ to perfectly model the intricate dependencies (the copula term $\mathcal{L}_c$) needed to cleanly isolate the $\ln(M^{(j)})$ terms in such cases. However, as observed in typical RL training states (e.g., Figure \ref{VDN-a estiamted} in the main text), target marginals are often adequately approximated as unimodal. Critically, the first two moments (mean and variance)—primarily required by our Forward SAC component—are often still well-captured by VDN-a estimations even with these complexities. Consequently, the VDN-a network provides a practically effective approach for approximating the necessary marginal properties within our Bidirectional SAC framework.

\section{More details of the Forward SAC algorithm}
\subsection{The implementation details of the Forward SAC algorithm}

\label{Integration_details}

The Forward SAC can be obtained from SAC by changing both the learning of the critic and actor in the following steps: for the critic, it uses the VDN-a network to serve as the Q-function, with the updating rule remaining the same as SAC; for the actor, Forward SAC uses the forward KL divergence and the explicit form of optimal projection policy, while the actor network learns directly from the mean and variance of the Boltzmann distribution $ q_i(\cdot \mid s_t) $.

In case the action space is typically bounded by using an invertible squashing function, such as $ \tanh(\cdot) $ \citep{SAC_arxiv, Distribution_SAC}, to squash the action samples from the Gaussian policy, the parameters for ultimate optimal projection policy is adapted as:
\begin{equation}
\label{final_tanh}
\begin{aligned}
    & f^{i}(s_t)^{*} = \int_{-b}^{b} q_i(\tanh(x) \mid s_t) \cdot x \cdot \tanh^{'}(x)\, dx, \\
    &\Sigma^{i}{}^{*} =  \int_{-b}^{b} q_i(\tanh(x) \mid s_t) \cdot (x - \mu)^2 \cdot \tanh^{'}(x) \, dx, \\
    & q_i(\cdot \mid s_t) = \frac{ \exp \left( \frac{1}{\alpha} Q_{i}(s_t, \cdot) \right) }{ \int_{-b}^{b} \exp \left( \frac{1}{\alpha} Q_{i}(s_t, \tanh(x)) \right) \tanh^{'}(x)\, dx }
\end{aligned}
\end{equation}
where $ \tanh^{'}(x) = (1 - \tanh^2(x)) $ is the derivative of $ \tanh(x) $, and $ b $ is a hyperparameter for the integration bound. The detailed derivation is described in Appendix \ref{tanh_s}. The numerical integration methods are adapted for the calculation of the optimal projection policy in Equation~\eqref{final_tanh}. We employ the widely recognized Simpson's integration method \citep{Simpson}. We denote the Simpson's integration as $\mathcal{S}_a^b[\cdot]$, which is renowned for its computational efficiency and high accuracy. For a given interval $[a, b]$, the definite integral of a function $g(x)$ is approximated by Simpson's rule as follows:
\begin{equation}
\label{Simpson_rule}
\int_a^b g(x)\, dx \approx \mathcal{S}_a^b[g(x)] = \frac{h}{3} \bigg[ g(x_0) + 4 \sum_{\substack{i=1 \\ i \text{ odd}}}^{I-1} g(x^i) + 2 \sum_{\substack{i=2 \\ i \text{ even}}}^{I-2} g(x^i) + g(x^i) \bigg],
\end{equation}
where the interval $[a, b]$ is partitioned into $I$ equally spaced subintervals of width $h = \frac{b - a}{I}$. The points $x_0 = a$ and $x^i = b$ denote the endpoints of the interval. The summations $\sum_{\text{odd}} g(x^i)$ and $\sum_{\text{even}} g(x^i)$ represent the aggregate of function evaluations at interior points with odd and even indices, respectively. 

Denoting the Simpson integration of the optimal projection policy as $ \hat{f}^{i}(s_t) $ and $ \hat{\Sigma}^{i} $, the Forward SAC algorithm can be implemented through two ways:

(1) Only use the critic. The mean and variance of the optimal projection policy $\pi_{\mathrm{new-f}}$ are calculated when interacting with the environment. This method ensures each iteration, the optimal projection policy is used.

(2) Use both the critic and actor. The actor network is updated through MSE loss, which is stable and efficient, but this actor interacting with the environment is not the exactly optimal projection policy (which is similar to the implementation of the Bidirectional SAC algorithm.

\begin{algorithm}
\caption{Forward SAC algorithm (only critic)}
\label{Forward SAC_only_critic}
\textbf{Hyperparameters:} Total number of training steps $\mathcal{L}$, number of time steps $\mathcal{T}$ each update epoch, mini-batch size $\mathcal{M}$, the update epoch $\mathcal{J}$ of neural networks  each training, and the learning rate $\beta$.
\begin{algorithmic}[1]  
    \STATE Initialize the critic network  $Q_{\theta}$ and replay buffer $\mathcal{B}$.
    \STATE Start with the initial state $s_0$
    \FOR{$l = 1, \dots ,\mathcal{L}$}
        \FOR{$t=1, \dots ,\mathcal{T}$}
        \STATE Calculate the mean $f^{i}(s_t)^{*}$ and variance $\Sigma^{i}{}^{*}$ in Equation \eqref{final_tanh} through Simpson's rule in \eqref{Simpson_rule}.
        \STATE Establish the optimal projection policy $\pi_{\mathrm{new-f}}$ through $f^{i}(s_t)^{*}$ and $\Sigma^{i}{}^{*}$.
        \STATE Sample actions from policy $\pi_{\mathrm{new-f}}$, collect and store the transition $(s_t, a_t, r_t, s_{t+1})$ in replay buffer $\mathcal{B}$.
        \FOR{$j=1,\dots,\mathcal{J}$}
        \STATE Sample mini-batch $\{(s_i, a^i, r_i, s_{i+1}) \mid i = 1, \dots, \mathcal{M}\}$ from $\mathcal{B}$.
        \STATE Update the critic network $Q_{\theta}$ by $\theta \leftarrow \theta-\beta \nabla_\theta J_Q(\theta)$ in \eqref{Q_updated}
        \ENDFOR
    \ENDFOR
    \ENDFOR
\end{algorithmic}
\end{algorithm}

For (1), the algorithm is detailed in Algorithm \ref{Forward SAC_only_critic}. For (2), the actor is updated by the gradient of the MSE loss function $ \nabla_{\phi} \text{MSE}( \pi_{\phi}) $:
\begin{equation}
\nonumber
\begin{aligned}
& \nabla \sum_{i=1}^{I}\left[\left(\hat{f}^{i}(s_t) - f_{\phi}^{i}(s_t)\right)^2 + \left(\hat{\Sigma}^{i} - \Sigma_{\phi}^{i}\right)^2\right] =\\
&2\sum_{i=1}^{I}\left[\left(\hat{f}^{i}(s_t) - f_{\phi}^{i}(s_t)\right)\nabla_{\phi} f_{\phi}^{i}(s_t)+ \left(\hat{\Sigma}^{i} - \Sigma_{\phi}^{i}\right)\nabla_{\phi}\Sigma_{\phi}^{i}\right].
\end{aligned}
\end{equation}
Then we show the algorithm of Forward SAC implemented as (2) in Algorithm \ref{Forward SAC}. Two soft Q-function networks and their corresponding target networks are adopted in this paper, similar to the SAC algorithm. The target network is updated smoothly by soft updates, where a hyperparameter $ \tau $ is used to control the updating speed.

\begin{algorithm}
\caption{Forward SAC algorithm (use an actor)}
\label{Forward SAC}
\textbf{Hyperparameters:} Total number of training steps $\mathcal{L}$, mini-batch size $\mathcal{M}$, the update epoch $\mathcal{J}$ of neural networks  each training, and the learning rate $\beta$
\begin{algorithmic}[1]  
    \STATE Initialize the critic network  $Q_{\theta}$, actor network $\pi_{\phi}$ and replay buffer $\mathcal{B}$.
    \STATE Start with the initial state $s_0$
    \FOR{$l = 1, \dots ,\mathcal{L}$}
        \STATE Using policy $\pi_{\phi}$, collect and store transitions $(s_t, a_t, r_t, s_{t+1})$ in replay buffer $\mathcal{B}$.
        \FOR{$j=1,\dots,\mathcal{J}$}
        \STATE Sample mini-batch $\{(s_i, a^i, r_i, s_{i+1}) \mid i = 1, \dots, \mathcal{M}\}$ from $\mathcal{B}$.
        \STATE Update the critic network $Q_{\theta}$ by $\theta \leftarrow \theta-\beta \nabla_\theta J_Q(\theta)$ in \eqref{Q_updated}
        \STATE Update the actor parameters with $\phi \leftarrow \phi-\eta \nabla_{\phi} MSE( \pi_{\phi})$. 
        \ENDFOR
    \ENDFOR
    
\end{algorithmic}
\end{algorithm}

\subsection{The relation between Forward SAC and variational policy optimization}
\label{Relation_VPO}
Variational policy optimization primarily relies on the EM method to iteratively update both the variational distribution and the policy. In the E-step \citep{MPO}, the variational distribution $ v(a_t \mid s_t) $ is learned iteratively, given by
\begin{equation}
    v(a_t \mid s_t) = \frac{\pi_{\phi}\left(a_t \mid s\right) \exp \left(\frac{1}{\alpha}Q_{\theta}(s, a_t)\right)}{\int_{\mathcal{A}} \pi_{\phi}\left(a \mid s\right) \exp \left(\frac{1}{\alpha}Q_{\theta}(s, a)\right)\, da}.
\end{equation}
This is similar to the Boltzmann distribution $ q(a_t \mid s_t) $. To update the policy, the M-step optimizes the ELBO with respect to $ \phi $:
\begin{equation}
\label{ELBO}
    \begin{aligned}
          \mathcal{J}(v, \phi) = \log p(\phi) + Q^{v}(s_t,\cdot)
          - \mathbb{E}_{\tau \sim v}\left[ \sum_{t=0}^{\infty} \left(  \alpha D_{\mathrm{KL}}\left( v(\cdot \mid s_t) \| \pi_\phi(\cdot \mid s_t) \right) \right) \right],
    \end{aligned}
\end{equation}
where $ p(\phi) $ is the prior distribution, and $ v(\cdot \mid s_t) $ is the variational distribution. If the prior distribution is set to the identity distribution, the optimization of the ELBO becomes equivalent to the optimization of the forward KL divergence, resembling our Forward SAC algorithm.

However, our Forward SAC algorithm differs from variational policy optimization in two key aspects:

\begin{enumerate}
    \item The Forward SAC algorithm uses the Q-learning process directly from the SAC algorithm, whereas variational policy optimization employs the E-step to optimize the variational distribution.
    \item The Forward SAC algorithm updates the policy by directly learning from the optimal projection, while variational policy optimization utilizes the M-step to optimize the ELBO for policy updating.
\end{enumerate}

Therefore, while there are similarities between our Forward SAC algorithm and variational policy optimization, the updating processes are fundamentally different.

\section{Broader impacts}
\label{Broader}
Our work on Bidirectional SAC presents several potential positive societal impacts. The improved sample efficiency could significantly reduce the computational resources required for training reinforcement learning agents, thereby lowering energy consumption and associated environmental impacts. Enhanced performance in continuous control tasks may accelerate advances in robotics for healthcare (e.g., assistive devices, surgical robots), industrial automation, and sustainable energy systems. The theoretical insights connecting forward and reverse KL divergence may contribute to the broader field of probabilistic machine learning.

\section{Experiment details}
\label{Effectiveness}
All environments tested in our experiments are trained for $10^6$ steps, except for the Pusher and Lunar Lander environments, which converge fast across all tested algorithms. To leverage GPU acceleration for training and simulation, we use the Brax simulator, which is built on Jax \citep{Brax}. All experiments are conducted on a cluster server equipped with an NVIDIA RTX A6000 GPU and 32 cores of an Intel(R) Xeon(R) Gold 5218 CPU operating at 2.30 GHz.

\subsection{More experiment results in episodic reward}
\label{More_results}
Here we demonstrate more experimental results in MuJoCo and Box2D environments.

\begin{figure}[htbp]
    \centering
    \includegraphics[width=\linewidth]{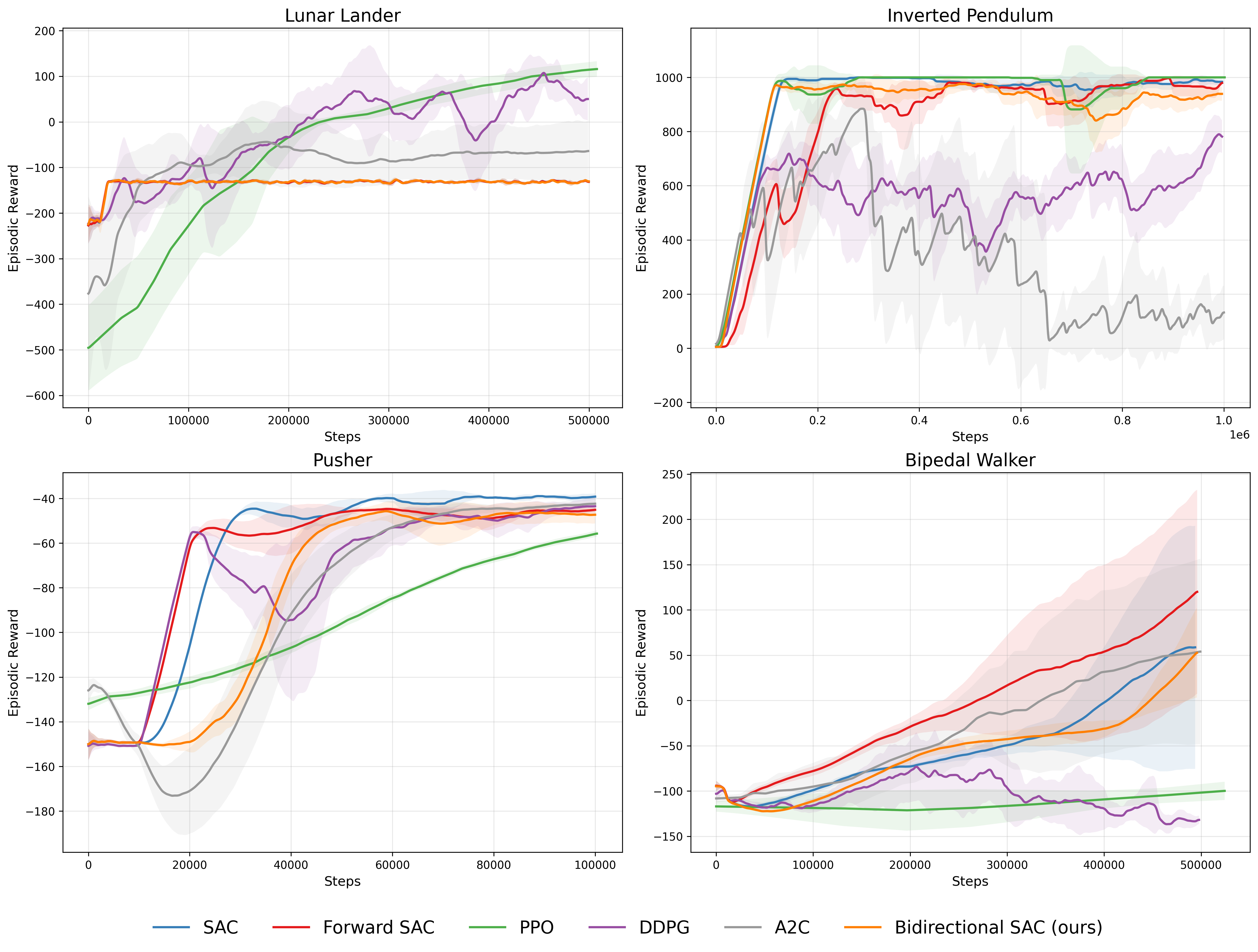}
    \caption{The episodic reward in MuJoCo and Box2D environments for Bidirectional SAC, Forward SAC, and other benchmarks in the training process.}
    \label{Compare_all_4}
\end{figure}
The experiment is further extended to more environments as shown in Figure \ref{Compare_all_4}. In the Lunar Lander environment, PPO exhibits superior performance by gradually achieving positive rewards, whereas both Bidirectional SAC and traditional SAC struggle to escape negative reward regions. DDPG demonstrates considerable variance but eventually reaches positive performance. For the Inverted Pendulum task, most algorithms converge rapidly to optimal performance, with traditional SAC, Bidirectional SAC, and PPO maintaining consistently high rewards. Forward SAC experiences some training instability before catching up, while A2C suffers significant performance degradation in later stages. Most notably, in the Bipedal Walker environment, Forward SAC distinguishes itself by attaining the overall best performance.

\subsection{More distribution comparison in update steps}
\label{dis_comparison}
\begin{figure}[htbp]
    \centering
    \includegraphics[width=\linewidth]{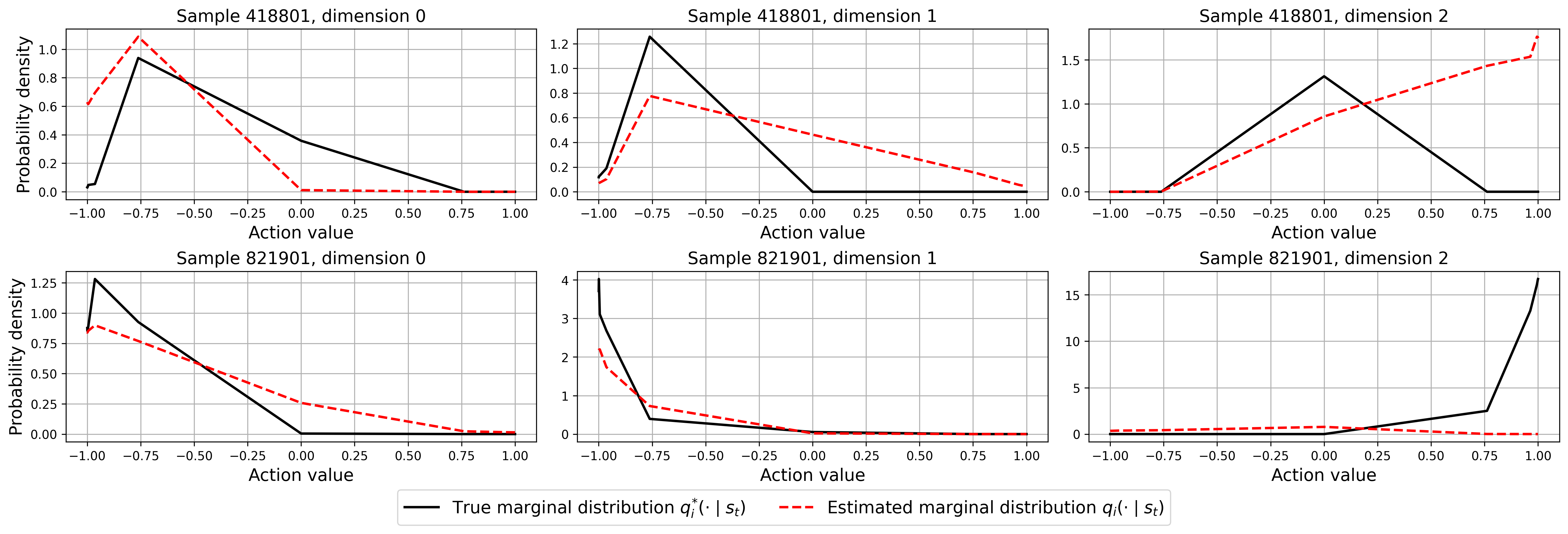}
    \label{dist_1}
\end{figure}

\begin{figure}[htbp]
    \centering
    \includegraphics[width=\linewidth]{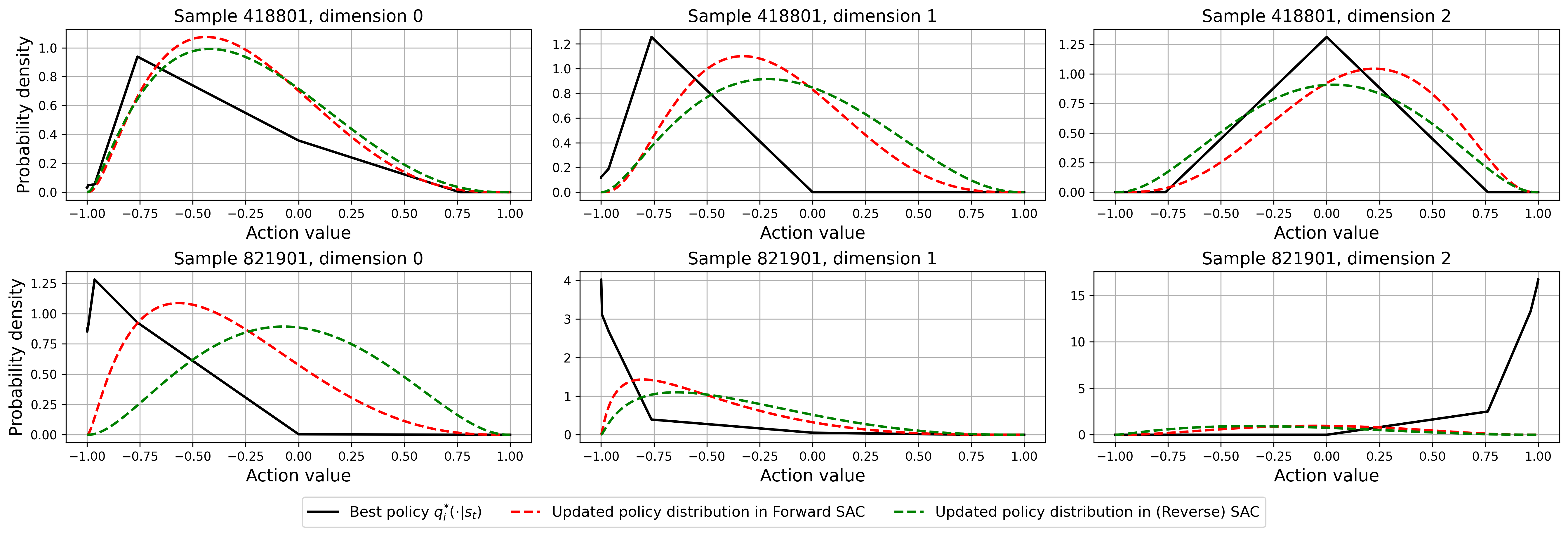}
    \label{dist_policy_1}
\end{figure}

\begin{figure}[htbp]
    \centering
    \includegraphics[width=\linewidth]{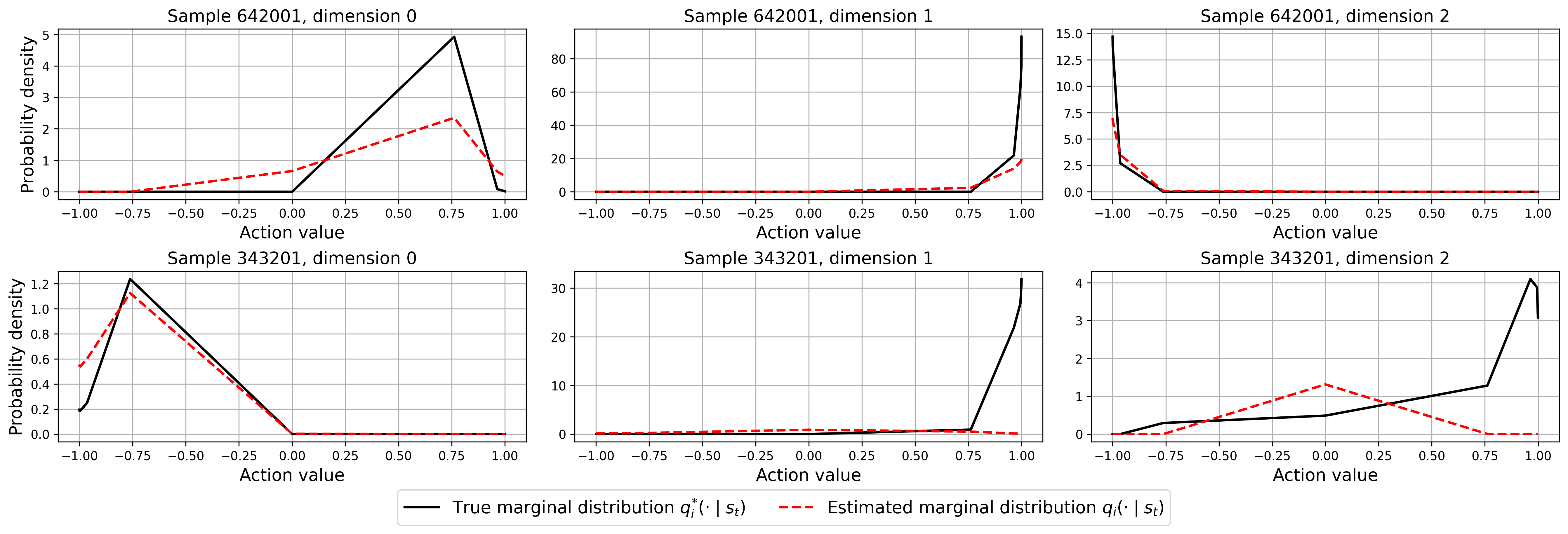}
    \label{dist_2}
\end{figure}

\begin{figure}[htbp]
    \centering
    \includegraphics[width=\linewidth]{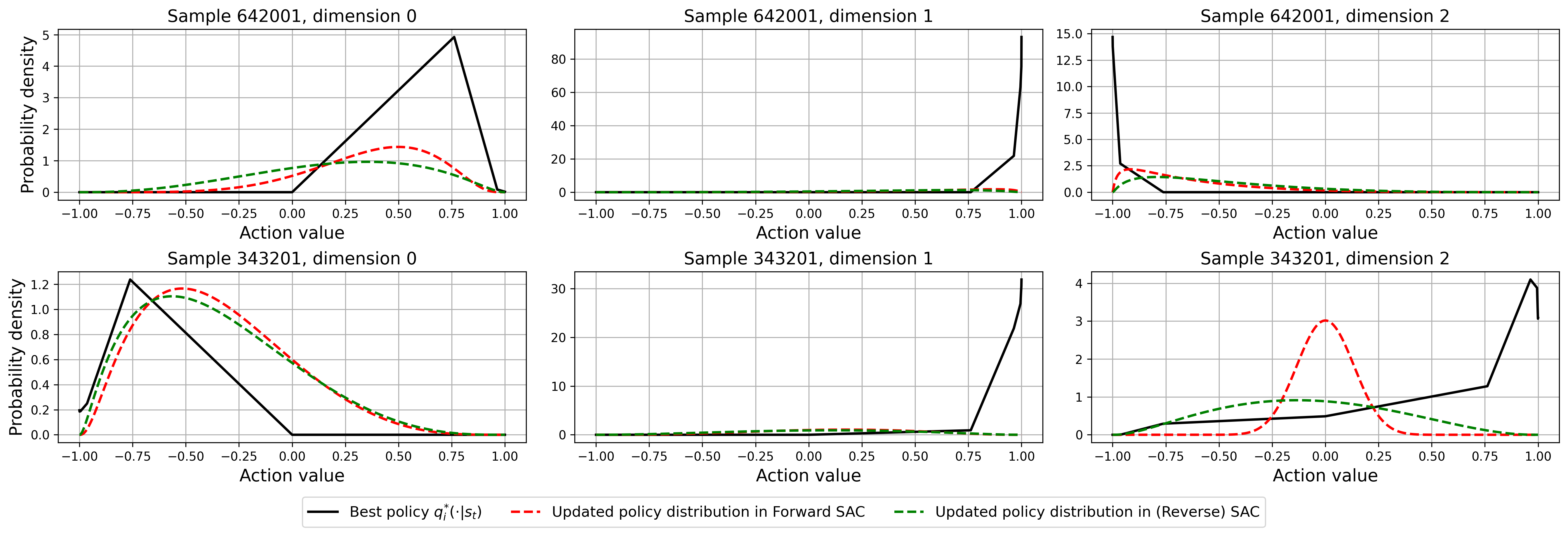}
    \label{dist_policy_2}
\end{figure}

\begin{figure}[htbp]
    \centering
    \includegraphics[width=\linewidth]{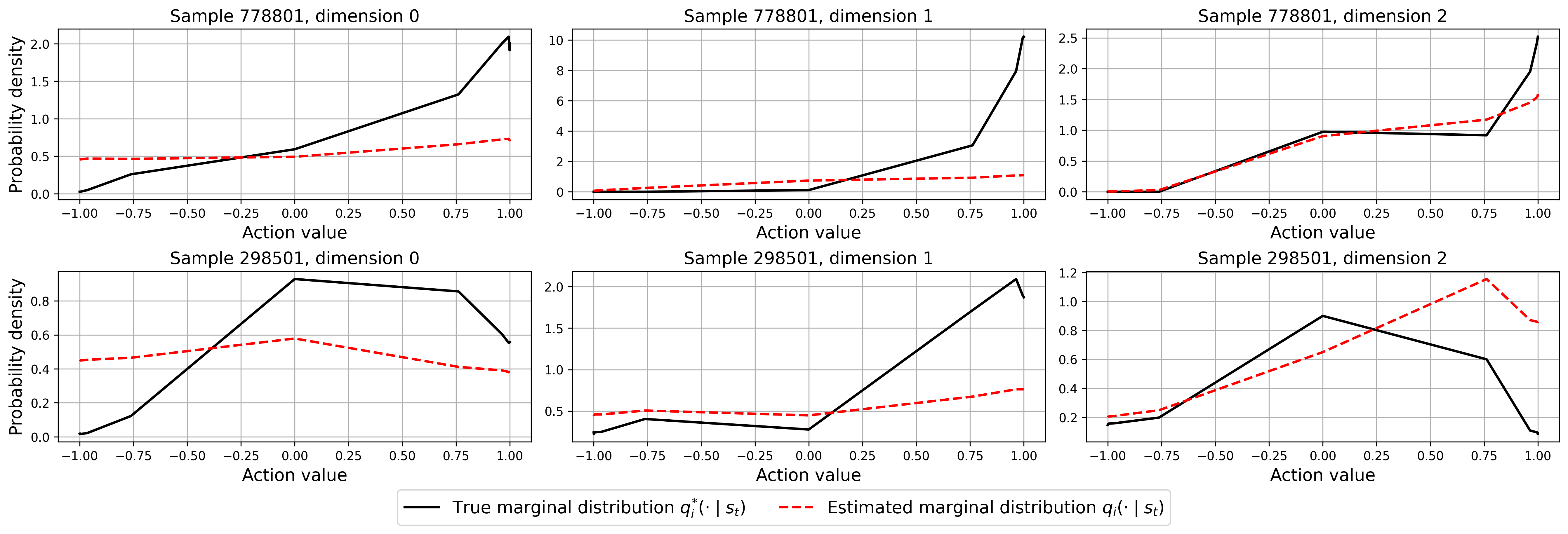}
    \label{dist_3}
\end{figure}

\begin{figure}[htbp]
    \centering
    \includegraphics[width=\linewidth]{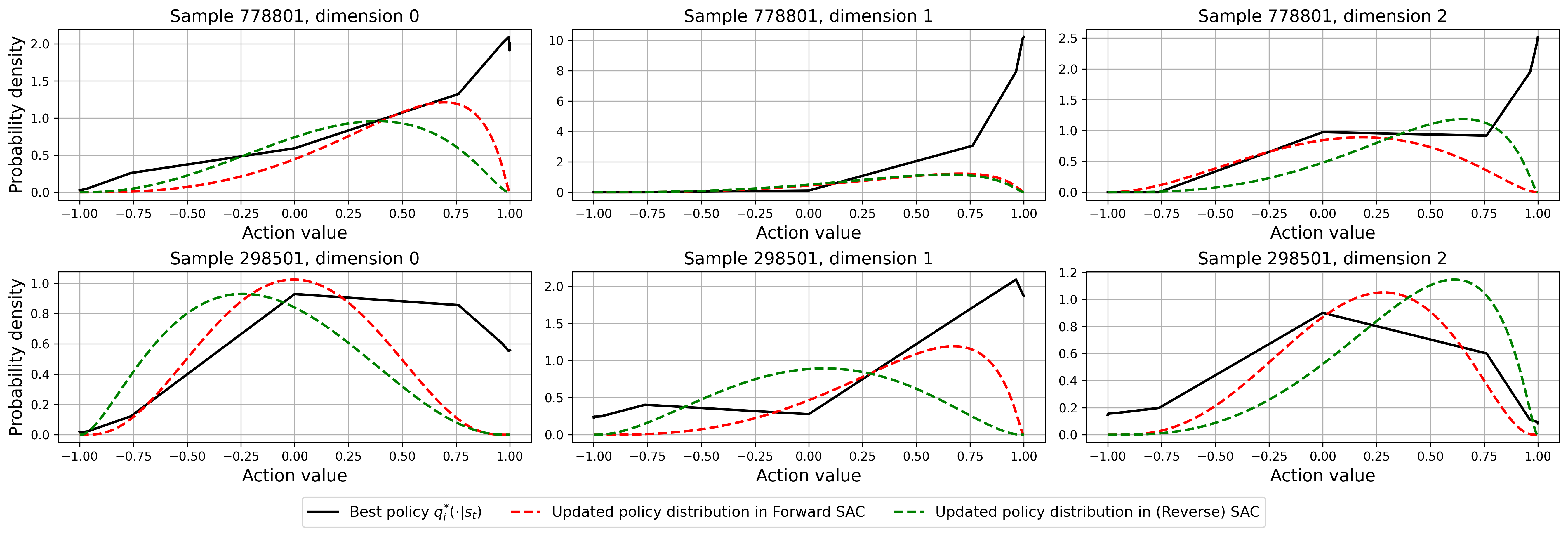}
    \label{dist_policy_3}
\end{figure}

\end{document}